\documentclass[10pt,twocolumn,letterpaper]{article}

\usepackage[pagenumbers]{iccv} 

\usepackage{multirow}
\usepackage{makecell}

\def\projname{RoboTron-Platter\xspace}
\def\pipelinename{RoboTron-Craft\xspace}
\def\srp{SRP\xspace}
\def\pip{PIP\xspace}

\definecolor{iccvblue}{rgb}{0.21,0.49,0.74}
\usepackage[pagebackref,breaklinks,colorlinks,allcolors=iccvblue]{hyperref}

\def\paperID{15489} 
\def\confName{ICCV}
\def\confYear{2025}

\title{Boosting Robotic Manipulation Generalization with Minimal Costly Data}

\author{
Liming Zheng
\and
Feng Yan
\and
Fanfan Liu
\and
Yufeng Zhong
\and
Chengjian Feng
\and
Lin Ma$^{*}$
}

\begin{document}
\maketitle
{\let\thefootnote\relax\footnotetext{Meituan Inc.}}
{\let\thefootnote\relax\footnotetext{$^{*}$ Corresponding authors.}}

\begin{abstract}
    The growing adoption of Vision-Language-Action (VLA) models in embodied AI intensifies the demand for diverse manipulation demonstrations.
    However, high costs associated with data collection often result in insufficient data coverage across all scenarios, which limits the performance of the models. 
    It is observed that the spatial reasoning phase (\srp) in large workspace dominates the failure cases. Fortunately, this data can be collected with low cost, underscoring the potential of leveraging inexpensive data to improve model performance. 
    In this paper, we introduce the \pipelinename, a  stage-divided and cost-effective pipeline for realistic manipulation generation. Base on this, the \projname method is introduced, a framework that decouples training trajectories into distinct task stages and leverages abundant easily collectible \srp data to enhance VLA model's generalization. 
    Through analysis we demonstrate that sub-task-specific training with additional \srp data with proper proportion can act as a performance catalyst for robot manipulation, maximizing the utilization of costly physical interaction phase (\pip) data.
    Experiments show that through introducing large proportion of cost-effective \srp trajectories into a limited set of \pip data, we can achieve a maximum improvement of 41\% on success rate in zero-shot scenes, while with the ability to transfer manipulation skill to novel targets.
    Project available at \url{https://github.com/notFoundThisPerson/RoboTron-Craft}.
\end{abstract}
\section{Introduction}
\label{sec:intro}

As the understanding and reasoning abilities of Multi-modal Large Language Models (MLLMs) advance rapidly, their application in real-world interactions, \ie Embodied Artificial Intelligence (EAI), has become a focal point of research~\cite{liu2024robomamba, brohan2023can, jin2024robotgpt}, and the method utilizing Vision-Language-Action (VLA) models is a common choice~\cite{kim2024openvla, zitkovich2023rt2, yan2024robomm, cheang2024gr2}. 
Similar to MLLMs, training the spatial understanding and physical interaction reasoning abilities of VLA requires a large quantity of demonstration trajectories across a variety of tasks. Although much effort at high cost has been dedicated to collecting robot demonstrations, both in simulation~\cite{gong2023arnold, mees2022calvin, ha2023scaling} and the real world~\cite{brohan2022rt1, rt-x, tan2024manibox}, generalizing agent-specific trajectories to a novel agent configuration remains a critical challenge. 
As a result, the training data available for a specified agent remains limited, which is far from sufficient to encompass the diverse real-world scenarios, thereby constraining the improvement of the VLA models' capabilities.

\begin{figure*}[t]
  \centering
  \includegraphics[width=0.75\linewidth, trim=50 20 50 20, clip]{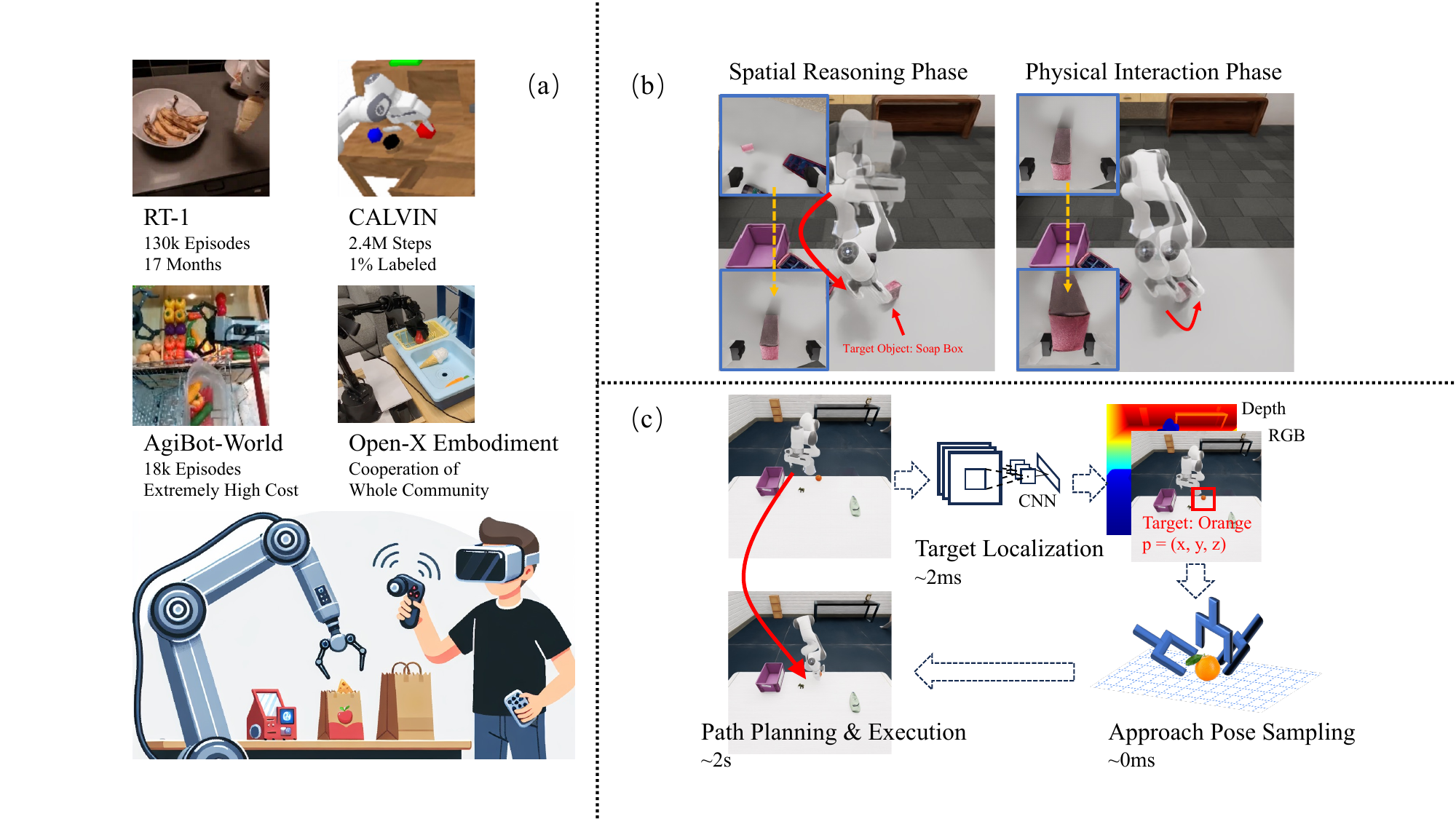}
  \caption{Motivation for Our Work: (a) Demonstration trajectories used in embodied AI training are typically collected via teleoperation, which is both time-consuming and expensive. (b) However, most task trajectories can be segmented into a spatial reasoning phase (\srp) 
  and a physical interaction phase (\pip), 
  each with distinct focus and learning difficulty. (c) The \srp data can be automatically collected using simple algorithms at a high speed.}
  \label{fig:motivation}
\end{figure*}

To address this issue and enhance data utilization efficiency, researchers are focusing on exploring cross-agent training~\cite{yan2024robomm, doshi2024scaling, rt-x, wang2025scaling, li2024towards}, spatial cognition enhancement~\cite{liu2024robouniview, huang2025enerverse, zhu2024spa} and task logical extraction~\cite{sermanet2024robovqa} through chain-of-thoughts.
Notably, recent studies~\cite{tan2024manibox, lin2024data} have demonstrated a scaling law governing the relationship between the spatial volume of operation workspace, the quantity of training data and the generalization performance of VLA models. All these approaches share a common premise: understanding the compositional nature of embodied tasks.

Through analyses we reveal that most tasks process can generally be divided into two stages: the Spatial Reasoning Phase (\srp) 
and the Physical Interaction Phase (\pip), 
as shown in \cref{fig:motivation}. 
The former stage is target-agnostic, as the agent explores extensive workspace without any close interaction with the targets, such as approaching the target before operation, making data collection relatively straightforward.
In contrast, during the later stage, precise actions governed by physical laws should be applied to the target with the foresight of object reaction, which is extremely labor-intensive, either for human or algorithmic experts. 
This motivates our core question: can inexpensive \srp data amplify the value of scarce \pip data thus reduce the effort required for data collection?

To answer these questions, we first introduce \pipelinename, a stage-divided and cost-effective pipeline designed to generate physically realistic manipulation trajectories with photo-realistic observation, along with an environment for model evaluation. By utilizing this pipeline, we produce diverse trajectories across various stages, encompassing a wide range of objects and environments. Based on which, we reveals the logarithmic scaling law between VLA model performance and target object diversity, amount of training trajectories, as well as environmental diversity.
Our key insight stems from two critical observations: (1) The spatial understanding ability required in \srp exhibits higher environmental variability compared to \pip, since the manipulation stage for a specified target is relatively fixed with little correlation with the surrounding scene; (2) Neural networks demonstrate distinct attention patterns during different task stages, such as the focus on target's location and spatial occupancy to avoid collision in \srp while shifted to the target proportion in \pip. These findings suggest that \textit{sub-task-specific training strategies} could better align with the model's learning characteristics, utilizing varying proportions of these data segments across the sub-tasks.

Both Tan \etal~\cite{tan2024manibox} and our experiments (see \cref{tab:approach_sr}) have demonstrated that a smaller workspace can significantly improve the success rate of operating tasks. This indicates that decoupling operation stages with different centers of attention can improve generalization performance. 
Furthermore, this variation in sub-task difficulty can cause the model to overfit on the simpler, small-workspace stage while underfit on the large-workspace stage, which necessitates different data volumes for each stage.
In this paper, we propose the \projname method, which decouples training data across different operation stages, constructs a implicit sub-task specific training procedure, and 
leverages a large amount of easily collectible \srp data to train this stage, to improve the performance of VLA models.
With our method, a substantial amount of labor-intensive teleoperation time traditionally required to collect complex manipulation trajectories, \eg the 17 months needed for 130k episodes in RT-1~\cite{brohan2022rt1}, can be significantly reduced.  Instead, program-driven automatic collection can be employed to acquire a large volume of low-interaction trajectories in extensive workspaces. This approach not only reduces manual efforts but also greatly enhances the potential to leverage larger datasets to improve model capabilities.

The contribution of this paper is as follows:
\begin{itemize}
    \item We introduce the \pipelinename, a stage-divided and cost-effective pipeline designed to generate realistic manipulation trajectories. Based on which we reveals the scaling law between model performance and trajectory diversity.
    \item We introduced the \projname methods, which utilizes additional cost-effective \srp trajectories to improve the model's generalization performance in zero-shot scenes.
    \item We prove that \srp data can act as catalyst to maximize the contribution of expensive manipulation dataset in VLA model training.
    \item Experiments demonstrate that our method increases the task success rate by 41\% in zero-shot scenarios and can effectively transfers model skills to novel target objects.
\end{itemize}


\section{Related Work}

\paragraph{Multi-modal VLA models} 
Unlike previous studies \cite{nair2022r3m, mees2023hulc2, zhao2023act} that employed models of limited size and do not heavily reliant on large volumes of training data, recent research efforts such as RoboMM~\cite{yan2024robomm}, RoboFlamingo~\cite{li2023vision} and $\pi_0$~\cite{black2410pi0} have leveraged MLLMs to achieve a generalist performance across multiple long-horizontal tasks through Imitation Learning (IL). Consequently, these approaches necessitate a substantial amount of data, imposing significant challenges in data collection. 
Numerous studies have invested considerable efforts in training with multiple datasets~\cite{rt-x, kim2024openvla, doshi2024scaling, wang2025scaling}. However, generalization across different tasks, embodiments, and datasets remains a significant challenge, necessitating further fine-tuning on specific datasets during evaluation. Another line of research utilizes pre-training with easily obtainable data formats~\cite{cheang2024gr2, liu2024robouniview, zitkovich2023rt2} to capture knowledge of the world, but still requires a large volume of action data to perform specified tasks effectively. Furthermore, diffusion-based methods~\cite{zhen20243d, ke20243d, li2024cogact, reuss2024multimodal, liu2024rdt}, as well as Vector-Quantization (VQ) methods~\cite{mete2025quest, szot2025grounding, lee2024behavior}, demand substantial amounts of action trajectories to adequately encapsulate high-dimensional probability distributions and codebooks. This paper proposes a data mixture method that reduces the reliance on costly manipulation trajectories, offering a partial solution to the aforementioned challenges.

\paragraph{Robot datasets \& benchmarks}
The EAI community have released a number of large-scale datasets collected in both simulation~\cite{gong2023arnold, mees2022calvin, zheng2024robocas, ha2023scaling, li2023behavior} and real world~\cite{rt-x, wu2024robomind, brohan2022rt1, walke2023bridgedata, fu2024mobile}. However, most datasets are collected through teleoperation and manual labeling, which is an extremely time-consuming process. Furthermore, the configurations of the embodiments, tasks and scenes in these datasets are different, posing challenges in reproducing performance in local experiments, particularly for datasets collected in real-world settings. On the other hand, datasets collected through algorithm-driven methods, which are primarily gathered in simulators using fixed task templates~\cite{zheng2024robocas, mandlekar2023mimicgen} or Reinforcement Learning (RL) with task disassembly~\cite{wang2023robogen, ha2023scaling, taomaniskill3}, are suffering from a lack of task diversity and often involve simplified physically simulations that are impractical for real-world deployment. With our method, models can be trained with a large proportion of easily collectible trajectories, which can be automatically collected through much simpler process, reducing the models' need for expensive interaction data.

\paragraph{Generalizing model capability}
Currently, most EAI models are limited to executing tasks they have explicitly encountered during training. For instance, even if a model be trained to pick up bottles, it cannot generalize this to pick up a cola can. Although this problem have already been studied through methods ranging from early domain randomization~\cite{james2019sim}, meta-learning~\cite{finn2017model} and data augmentation\cite{laskin2020reinforcement} to recent advancements in world model building~\cite{cheang2024gr2, liu2024robomamba} and spatial reasoning~\cite{liu2024robouniview, huang2025enerverse}, the generalization performance on out-of-distribution (OOD) novel targets still shows limited improvement. \cite{cheang2024gr2, zitkovich2023rt2} try to transfer the world knowledge from large models trained with Internet-scale data to robot action reasoning, but the the manipulation experience of OOD targets from ``practicing'' can not be efficiently acquired from ``reading'', while \cite{liu2024robomamba, sermanet2024robovqa, sermanet2024robovqa} are trying to directly use the general ability to guide the agent's action logic. \cite{liu2024robouniview, huang2025enerverse, zhu2024spa} are working improving the action performance through understanding the spatial information in the workspace. Zhu \etal~\cite{zhu2025objectvla} transfer the target knowledge to similar objects through text-image pairs, but still needs auxiliary information to get a better performance during inference. In this paper we propose an end-to-end training method, which can improve the generalization performance on OOD targets  by a large margin.
\section{\pipelinename Benchmark}

\begin{figure}
    \centering
    \includegraphics[width=0.95\linewidth, trim=250 0 250 0, clip]{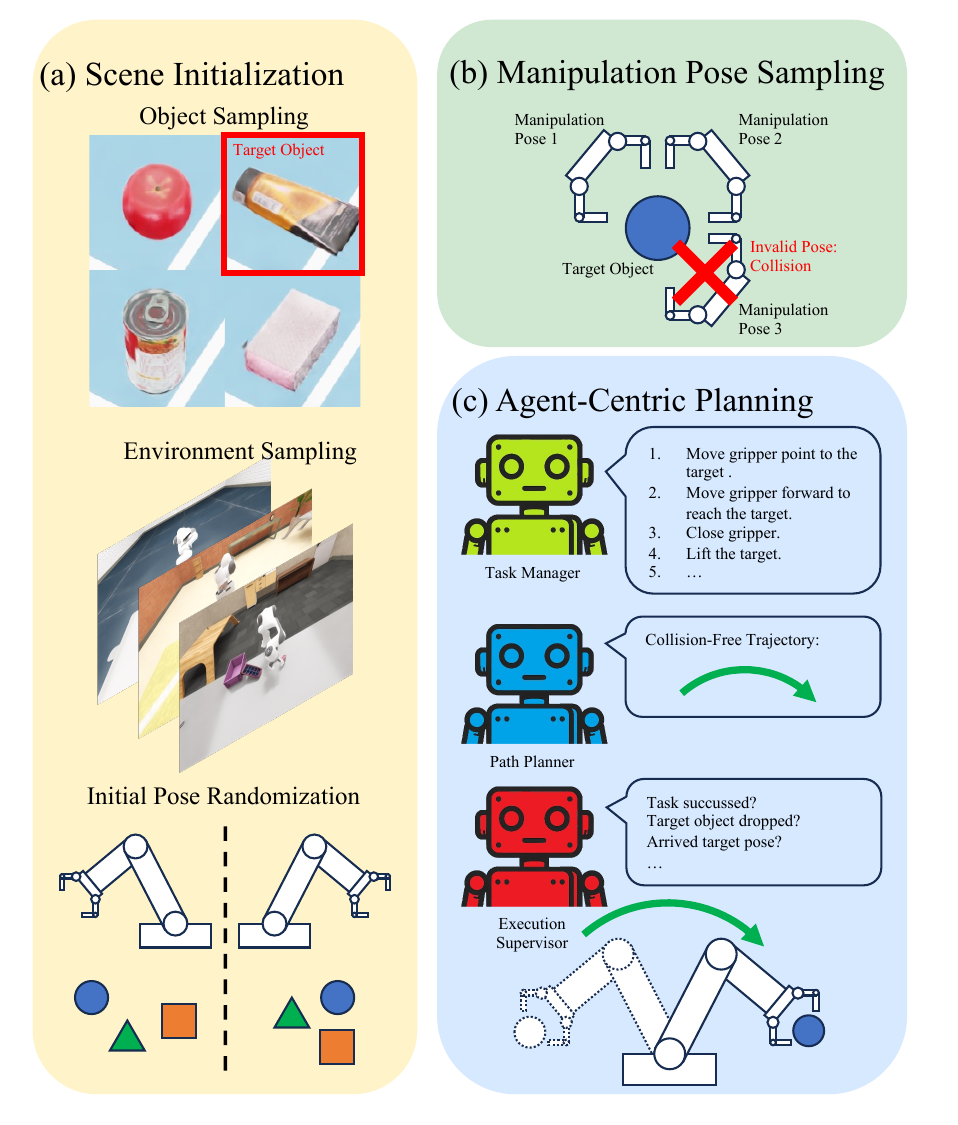}
    \caption{Overview of our data generation pipeline. (a) The environment is first initialized with multiple objects with random pose. (b) In the object-centric planning phase, feasible manipulation pose of the target is sampled. (c) In the agent-centric planning phase, our pipeline split the task into several segments, plan the execution trajectory and watch the task progress during execution.}
    \label{fig:data-pipeline}
\end{figure}

To emphasize the ability of behavior cloning methods in manipulating various shaped objects in real world, a substantial number of demonstration trajectories is essential. Accordingly, as shown in \cref{fig:data-pipeline}, we introduce the \pipelinename benchmark, which offers a cost-effective pipeline for generating physically realistic manipulation trajectories, along with an environment for model evaluation.
Built upon the powerful NVIDIA Isaac Sim~\cite{isaacsim} platform, our benchmark can provide a smooth robot trajectory generation and evaluation pipeline on various commonplace objects, with photo-realistic rendering in multiple well-appointed scenes and reasonable physical feedback on interaction-rich tasks. Details are shown in \cref{sec:data-benchmark}.

\subsection{Trajectory Generation Pipeline}
We automatically generate robot demonstration trajectories on manipulation tasks with target objects in various geometry shapes, scattered on several workbenches in different rooms with various background textures. 
The generation pipeline is decomposed into a two-level hierarchy: object-centric planning and agent-centric planning, for the convenience of scalability across various task types.

Starting from randomly sampling a pre-defined environment and several manipulatable objects $\left\{ \mathcal{M} \right\}$, our benchmark randomly places the objects in the scene with collision-free poses. In the following object-centric planning phase, our benchmark samples a target object $\mathcal{M}_\text{tgt} \in \left\{ \mathcal{M} \right\}$, along with a feasible manipulation pose $\mathbf{g} = \left(\mathbf{R}, \mathbf{t}\right) \in SE(3)$ of $\mathcal{M}_\text{tgt}$, which is sampled from a set of pose candidates $\mathcal{G}^\text{cand}_\mathcal{M}$ that are automatically generated using method described in \cref{sec:label-annotation}. Then during the agent-centric planning, task execution stages are supervised by the task manager with the known $\mathbf{g}$ and task template, and the collision-free robot trajectories in each stage are planned using motion planners with the priori knowledge about the geometry shape of the target and the spatial occupancy of the scene. Details can be found in \cref{sec:traj-gen}.

\subsection{\srp Data Collection}
\label{sec:srp-data-collection}
Although it is efficient to collect demonstrations through our \pipelinename pipeline in simulation, it is not implementable in real robot data collection, which is necessary to bridge the sim-to-real domain gap. On the other hand, during trajectory generation, time consumed by the motion planning when approaching the target and contact simulation when manipulating the target takes over 70\% of the progress, eliminating these cost can greatly speed up data collection. Luckily, our experiment shows that enriching the spacial reasoning stage trajectories can promote scene generalization, meaning that we can collect the trajectory slice that only approaching the target without actual interaction.

To achieve this, starting from a ground truth manipulation pose sample $\mathbf{g} \in \mathcal{G}^\text{cand}_\mathcal{M}$ of target $\mathcal{M}_\text{tgt}$, an offset transform $\Delta \mathbf{T} \in SE(3)$, with random orientation offset $\Delta \mathbf{R}$ and translation offset $\Delta \mathbf{t}$ away from the object, is sampled and applied to $\mathbf{g}$, \ie
\begin{equation}
    \mathbf{g}_\text{app} = \mathbf{g} \Delta \mathbf{T},
\end{equation}
to get the approaching pose $\mathbf{g}_\text{app}$ just before interacting with $\mathcal{M}_\text{tgt}$, which serves as the final state of our \srp trajectories.
In this way, the need for planning the narrow path reaching exactly the target object for the motion planner as well as the complex collision computation of the physics simulator while interacting can be eliminated, which greatly saves the data collection time.
Since the manipulation model can learn the correct interaction pose with the object, it will automatically fix the pose error while inference as long as the pose error is within a certain range, which is used as the sample space of $\Delta \mathbf{T}$. The parameters used in this stage are shown in \cref{sec:data-parameters}.

Further more, this procedure can be applied to the real-world robot trajectory collection. With sparse ground truth manipulation labels on the target object and an object pose estimation algorithm, \srp trajectory collection can be greatly accelerated since the robot can plan and execute the trajectories automatically, reducing the need for human tele-operation, thus the data richness can be greatly improved, providing the potential for better capability of VLA models in manipulation.

\section{\projname: Training with Data Mixture}
\label{sec:data-mixture}

\begin{figure*}[t]
  \centering
  \includegraphics[width=0.9\linewidth, trim=0 20 0 20, clip]{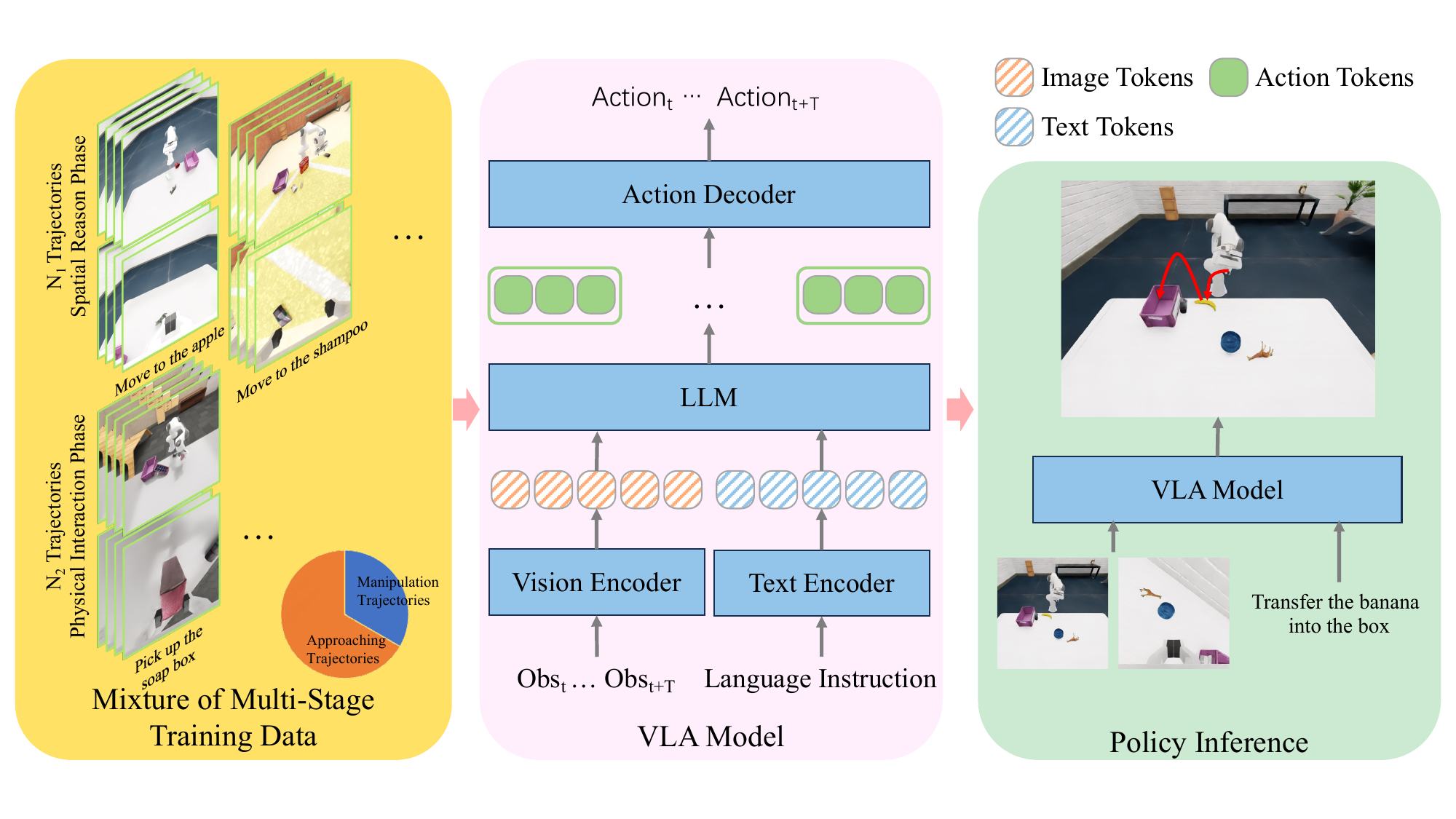}
  \caption{Overview of our method. We divide each training trajectory into two stages: the spatial reasoning phase (\srp), which does not require close interaction, and the physical interaction phase (\pip), during which the agent directly operates the target. $N_1$ \srp and $N_2$ \pip trajectories are sampled to form a new dataset (left). This resampled dataset is used to train the VLA model in an IL manner (middle). During the inference process, the VLA model takes the agent's observation and language instruction as input and predicts the next step action to guide the agent in accomplishing the task (right).}
  \label{fig:overview}
\end{figure*}

As illustrated in \cref{fig:overview}, this paper introduces the \projname method, which segments robot manipulation trajectories into spatial reasoning and physical interaction phases according to the extent of the agent's interaction with objects in the environment. By employing a mixture of two-stage data in appropriate proportions, we aim to achieve a generalization performance comparable to using complete data for model training. This approach can effectively reduce the reliance on expensive \pip data.

In order to train the VLA model, we first need to collect a dataset of full-stage demonstration trajectories $\mathcal{D}^F = \left\{ \tau^F_i \right\}_{i=1}^{N}$ for supervised training. Note that we are aiming to train the VLA model with a larger quantity of easily-collectible \srp data than expensive \pip data, so independently-collected \srp dataset $\mathcal{D}_{ind}^{\srp}$ can be included in our training process.


To utilize the trajectories of different sub-tasks in both $\mathcal{D}^F$ and $\mathcal{D}_{ind}^{\srp}$, we first segment the given full-stage trajectory $\tau_i^F \in \mathcal{D}^F$ into \srp and \pip based on the distance between the end-effector $G$ and the target object $T$. 
More formally, the \pip segment $\tau^{\pip}_i$ starts from the state that $G$ is at the approaching pose $\mathbf{g}_{\text{app}}$, and ends up when the interaction-rich stage is accomplished, \eg after grasping the target in pick-and-place task or triggering the button in switch-operation task. The remaining segments of $\tau_i^F$ such as the approaching stage are all classified into \srp segments $\tau^{\srp}_i$.
Following such procedure, the trajectory can be divided into several segments $\tau_i^F = \left\{ \tau_{i, 1}^{\srp}, \tau_{i, 1}^{\pip}, \tau_{i, 2}^{\srp}, \cdots \right\}$. Correspondingly, the dataset can be divided into two sub-datasets: $\mathcal{D}^F = \mathcal{D}^{\srp} \cup \mathcal{D}^{\pip}$, 
where $\mathcal{D}^{\srp} = \left\{ \tau_{i, j}^{\srp} \right\}$ and $\mathcal{D}^{\pip} = \left\{ \tau_{i, j}^{\pip} \right\}$.

Before the training phase of the VLA model, we sample $N_1$ and $N_2$ segments in $\mathcal{D}^F$ and $\mathcal{D}_{ind}^{\srp}$ respectively, and construct a new dataset $\mathcal{D}^{Mix}$ to train the model, which in this paper we call \projname, \ie 
\begin{equation}
    \mathcal{D}^{Mix} = \left\{ \tau_i^F \sim \mathcal{D}^F \right\}_{i=1}^{N_1} \cup \left\{ \tau_i^{\srp} \sim \mathcal{D}_{ind}^{\srp} \right\}_{i=1}^{N_2}.
    \label{equ:mix-dataset}
\end{equation}
In practice, to achieve the best model capability, generally the whole full-stage trajectory dataset $\mathcal{D}^F$ is used, \ie $N_1 = \left|\mathcal{D}^F\right|$, and select a proper $N_2$ to improve the generalization performance on novel scenes.
Through this method, a implicit sub-target specific training with sub-task datasets $\mathcal{D}^{\pip}$ and $\mathcal{D}^{\srp} \cup \mathcal{D}_{ind}^{\srp}$ is constructed, providing a flexible way to control the performance of each sub-task.
By varying the proportion of data between the two sub-datasets, a tendency in the task success rate relative to the amount of \srp data can be observed in \cref{sec:exps}, from which a principle for conserving \pip data while maintaining the VLA model's performance can be concluded.
\section{Experiments}
\label{sec:exps}

In this section, we investigate how the total number of training trajectories and the proportion of \srp data impact task success rates. We aim to identify the optimal strategy for leveraging easily accessible \srp data to enhance the generalization performance of the VLA model.
Moreover, we investigate the scaling law of training data amount and the diversity of target geometries as well as environment setups. 

\subsection{Environment Setup}

Unless otherwise stated, in this paper the RoboMM~\cite{yan2024robomm} is used as our baseline. Language instruction, RGB images from wrist and table cameras, and camera parameters are feed into the model in the training phase.
During training we form the mixed dataset $\mathcal{D}^{Mix}$ with a varying of proportion of independent \srp segments, to verify the effectiveness of our \projname method, \ie
\begin{equation}
    p_{\srp} = \frac{N_2}{N_1 + N_2}.
\end{equation}
In this paper we use trajectories pick-and-place task on different target objects as our training data.
The trained models are evaluated in our \pipelinename benchmark using IsaacSim platform. A trail is considered succeeded if the agent successfully picked up the instruction-specified target object under the actions generated by the VLA model. 
Note that depth images are not utilized in evaluation, which is only used in training for supervision together with action trunks.
More details are shown in \cref{sec:env-setup}.

\subsection{Experiment Results for \projname}
\subsubsection{Generalization with \srp Data}

\begin{figure}[t]
  \centering
  \includegraphics[width=\linewidth, trim=20 110 20 110, clip]{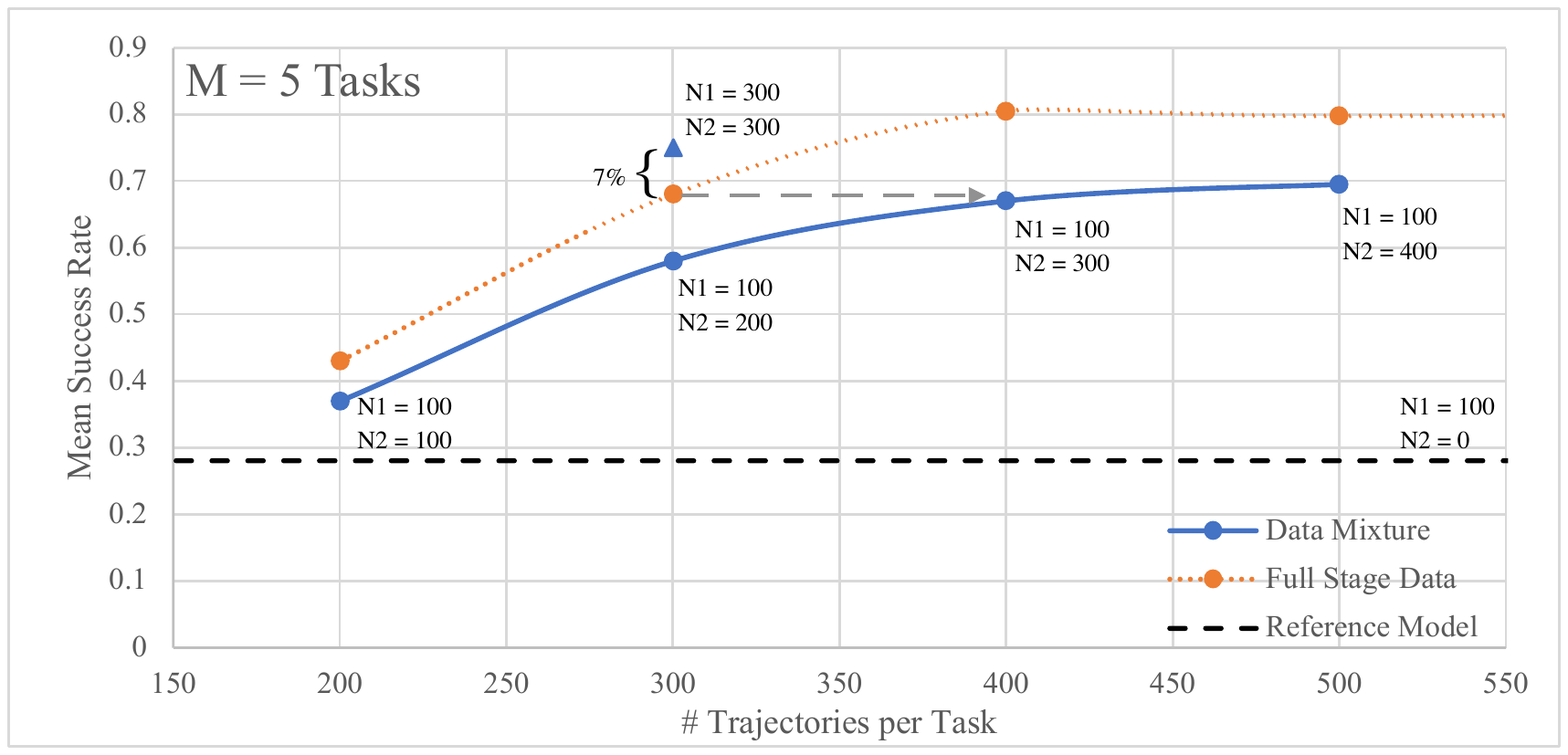} \\
  \includegraphics[width=\linewidth, trim=20 110 20 110, clip]{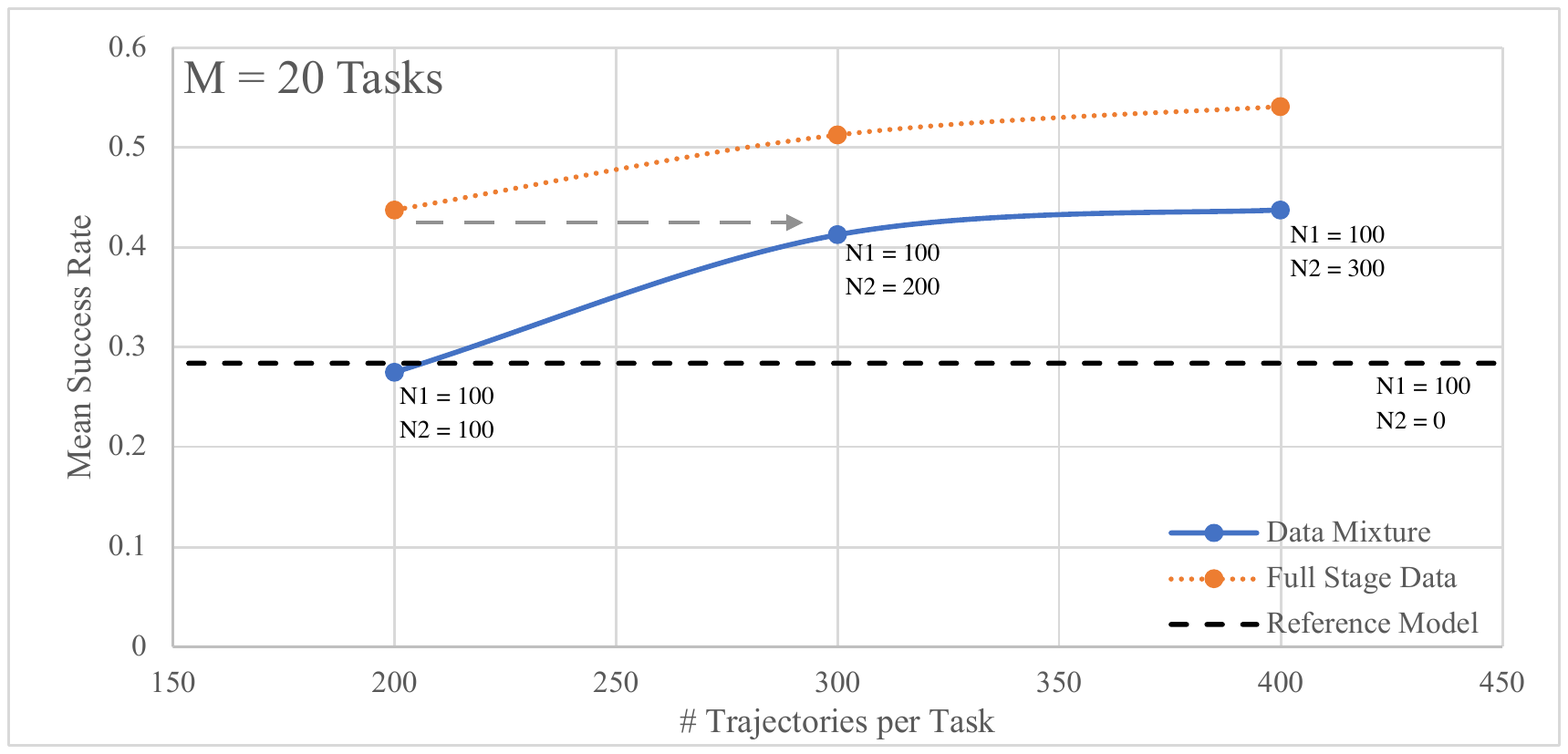}
  \caption{Impact of training trajectories quantity per task on mean success rate among all tasks in zero-shot scenes. 
  The reference model is trained using with a configuration of $N_1=100, N_2=0$ to serve as the baseline performance.
  The horizontal axis represents the number of training trajectories for a single task, \ie $N_1 + N_2$.
  }
  \label{fig:sr_wrt_trajs}
\end{figure}

Firstly, we aim to determine whether increasing the amount of \srp data can improve the model's generalization performance. 
We randomly select the \pip trajectories of $M$ target objects as our task, and  create a multi-task dataset to train the RoboMM models. For each task, we use a fixed number of $N_1=100$ full-stage trajectories $\tau_i^F \in \mathcal{D}^F$, alongside a variable number of $N_2$ \srp trajectories $\tau_i^{\srp} \in \mathcal{D}_{ind}^{\srp}$. The results are presented in \cref{fig:sr_wrt_trajs}. Models trained with totally full-stage (FS) data ($N_2=0$) are also presented as a reference, to serve as the upper bound of model performance.

With a maximum margin of 41\% over the reference model, incorporating additional automatically collected \srp data during training significantly elevates the model's success rate, and can achieve an equivalent performance (marked with gray arrow) to that trained with high-cost full-stage data.
However, the performance bottleneck occurs earlier than that of using full-stage data, when $N_2 > 2 N_1$ ($N_2 > 200$ in this experiment), 
increasing $N_2$ while keeping $N_1$ constant yields no significant improvement in the model's performance. 
This suggests that incomplete trajectories cannot be added indefinitely, as insufficient operational data may hinder the model from learning effective manipulation ability, and the proportion of data mixture is discussed in \cref{sec:proportion}. 
Increasing $N_1$ can delay the arrival of the bottleneck, as another $7\%$ of promotion is achieved through increasing $N_1$ from 100 to 300, which is close to the bottleneck achievable with full-stage data.

\subsubsection{Universality of \srp Data}
To verify that adding \srp data can improve the model performance across different models and datasets, we trained and tested RoboMM~\cite{yan2024robomm} and RoboFlamingo~\cite{li2023vision} on both our dataset and CALVIN dataset~\cite{mees2022calvin} using the aforementioned approach. 

As shown in \cref{tab:diff_models}, in all of the settings, adding additional \srp segments data can significantly improve the generalization performance in different models and different tasks. 
The improvement amplitude compared to the baseline model (marked ``w/o \srp'') on our dataset is significantly higher than on CALVIN. This is because the tasks in CALVIN are simpler, while the trajectories approaching the target in our dataset are more complex. In our dataset, agents approach the target from random orientations rather than a relatively fixed pose, as in CALVIN, resulting in a larger search space for the \srp policy. These findings underscore the advantage of our method in handling tasks with expansive workspaces.

\begin{table}
  \centering
  \begin{tabular}{@{}ll|cc@{}}
    \toprule
    Dataset & Model & w/o \srp & w/ \srp \\
    \midrule
    \multirow{2}{*}{Ours} & RoboMM & 0.28 & 0.58 \\
    & RoboFlamingo & 0.13 & 0.28 \\
    \midrule
    \multirow{2}{*}{CALVIN} & RoboMM & 0.80 & 0.88 \\
    & RoboFlamingo & 0.74 & 0.78 \\
    \bottomrule
  \end{tabular}
  \caption{Model performance with additional \srp segments on multiple models and datasets. w/ \srp: models trained with \srp data at $p_{\srp} = 66\%$;  w/o \srp: models trained with the same $N_1$ while $N_2=0$.}
  \label{tab:diff_models}
\end{table}

\subsubsection{Generalization on Novel Target}
Additionally, for tasks requiring a deep understanding of object geometry and physical laws, such as picking up objects with totally different geometries, we wonder whether the VLA models can transfer generalized knowledge to out-of-distribution (OOD) target objects. 
To this end, we introduce new tasks featuring novel target objects and only \srp data during training, within datasets containing other targets with different geometries and totally full-stage trajectories, and see that if the novel task can be successfully executed.
The results are shown in \cref{tab:half_trajs}. 

With only \srp data, the models can successfully pick up the novel target even if they have not seen examples on how to do this, especially for the 5 tasks with similar geometry (see \cref{fig:objs}), only through the experience on the other targets. Meanwhile, without \srp segments, the models are just wandering in the workspace without knowing what to do. 
This result indicates that the similarity between the \srp segments of the novel targets and the others acts as a bridge, enabling the models expand the skill of the entire task to the novel target, without requiring any additional auxiliary information. This finding significantly broadens the scope for future research on task generalization performance.
However, performance on novel targets remains relatively low, indicating that further research is needed in this area.

\begin{table}[tp]
  \centering
  \begin{tabular}{@{}ll|cc@{}}
    \toprule
    \# Tasks & Config. & Seen Targets & Novel Targets \\
    \midrule
    \multirow{3}{*}{\makecell[l]{4 Seen\\\&1 Novel}} & None & 0.67 & 0.05 \\
    & \srp & 0.65 & 0.40 \\
    & \srp + \pip & 0.64 & 0.65 \\
    \midrule
    \multirow{3}{*}{\makecell[l]{10 Seen\\\&10 Novel}} & None & 0.62 & 0.03 \\
    & \srp & 0.61 & 0.20 \\
    & \srp + \pip & 0.74 & 0.65 \\
    \bottomrule
  \end{tabular}
  \caption{
    Generalization performance on OOD target objects. Performance of in-distribution targets are also presented as a control. Data configuration: ``None'' - No training data on novel target objects; ``Only \srp'' - Novel target objects have only \srp segments in the dataset; ``\srp + \pip'' - Novel objects are trained using the same full-stage data as other tasks.
  }
  \label{tab:half_trajs}
\end{table}



\subsection{Data Scaling Law}
To verify whether there is a scaling law in the performance of manipulation tasks under conditions of target geometric diversity and environmental diversity, we trained models using the pick-and-place task under various settings and evaluated their performance in novel scenes. In each trial, a score of 1 is given if the target object is successfully picked up and placed into a box; a score of 0.5 is assigned if the target is dropped during the transfer process; otherwise, the trial receives a score of 0. The mean score for each model is reported in this paper.

\subsubsection{Object Instance Capability}
\begin{figure}[tp]
    \centering
    \includegraphics[width=0.95\linewidth, trim=2 2 2 2, clip]{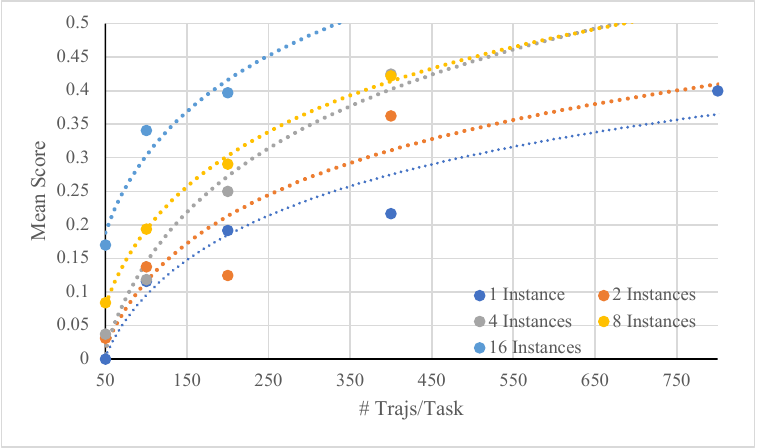}
    \caption{Performance of target instance scaling in zero-shot scenes. The horizontal axis shows the number of training trajectories per object used for each model.Each curve represents the performance of a series of models trained using different numbers of target instances.}
    \label{fig:multi-task-result}
\end{figure}

To explore the scaling low between target geometry diversity and the amount of demonstrations, experiments shown in \cref{fig:multi-task-result} are performed. Note that in each scene there are 3-5 interfering object except the target.


The results reveal several notable patterns: (1) Increasing the quantity of training data for a single instance, as well as augmenting the number of training instances with different geometries within a single model, can enhance the policy's performance. This indicates that the agents possess the capability to manipulate diverse-shaped target objects during real-world deployment. (2) Instance capability improvement can be more effectively achieved by augmenting the diversity of target geometries rather than by merely enriching the quantity of training trajectories for each instance, which suggests that the policy can abstract geometric feature that are suitable for manipulation from diverse-shaped targets. 
Furthermore, this capability allows VLA models to handle objects with unseen shapes by utilizing a limited set of training objects that cover sufficient combinations of geometric parts. 



\subsubsection{Environment Capability}

\begin{figure}[tp]
    \centering
    \includegraphics[width=0.95\linewidth, trim=2 2 2 2, clip]{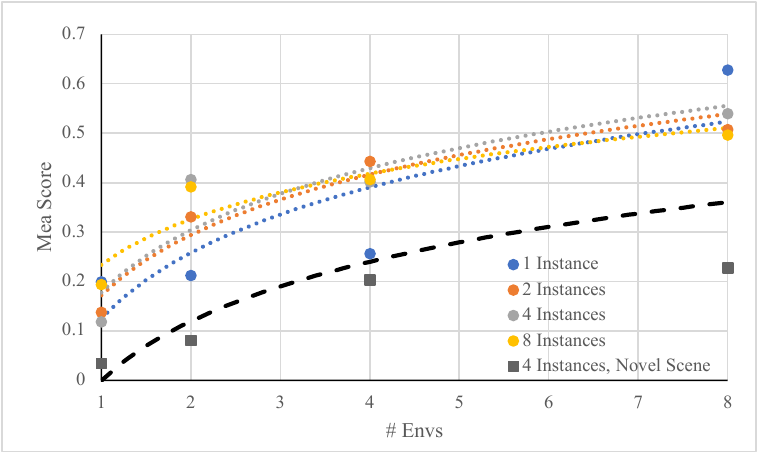}
    \caption{Performance of the policies trained on different quantity of environments on test scenes. Each task use 100 demonstration trajectories in each scene.}
    \label{fig:multi-env-result}
\end{figure}

To determine whether the VLA model can generalize to novel environments, we trained several models using 100 trajectories for each target object in various environments, with a varying number of scene backgrounds and target objects. The results, illustrated in \cref{fig:multi-env-result}, indicate that the performance of the VLA model positively correlates with the diversity of training scenes. This correlation is observed not only in environments encountered during training but also in new environments, suggesting that diversity in training environments is crucial for enhancing the model's generalization capability.
However, increasing the diversity of target object instances does not significantly improve model performance, especially after expanding the number of training environments. This indicates that enhancing environmental diversity provides greater benefits than increasing object diversity for VLA models.


\subsection{Ablation Studies}
\label{sec:ablation}

\subsubsection{Ablation in Demonstration Generation}

To investigate the attributes that constitute high-quality data and to facilitate the development of policies for enhanced generalization performance, we omitted specific features during the data generation process and assessed the models trained on this modified dataset. The results are presented in \cref{tab:ablation}.

\begin{table*}[tp]
    \centering
    \begin{tabular}{ll|c}
        \hline
        \multicolumn{2}{c|}{Settings} & Performance\\
        \hline
        \multirow{3}{*}{\makecell[l]{Approaching\\Target}} & Random Direction Approach & 0.3 \\
        & Strict Linear Approach & 0.15 \\
        & Relaxed Linear Approach & \textbf{0.55} \\
        \hline
        \multirow{3}{*}{Randomization} & \multirow{2}{*}{Fixed Manipulation Pose} & 0.45 (On Object Pose with Random Yaw) \\
        & & 0.025 (On Randomized Object Pose) \\
        & Randomized Manipulation Pose & \textbf{0.35} \\
        \hline
        \multirow{2}{*}{\makecell[l]{Action Label\\Generation}} & Target Pose Subtraction & 0.075 \\
        & State Subtraction & \textbf{0.35} \\
        \hline
        \multirow{2}{*}{\makecell[l]{Occlusion in \\ Observation}} & Head Camera + Wrist Camera & 0.55 \\
        & Table Camera + Wrist Camera & \textbf{0.625} \\
        \hline
    \end{tabular}
    \caption{Performance of models trained on datasets generated with distinct features.}
    \label{tab:ablation}
\end{table*}

At the stage before manipulate the target object, it is beneficial to approach the target with a relaxed linear trajectory, \ie move the gripper and wrist camera towards the target object while maintaining a certain distance first, then transition to the manipulation pose following a nearly linear trajectory. 
Through such a path, the observation problem of wrist camera, caused by approaching from random directions, can be avoided. Additionally, the rigidity associated with strictly linear paths in the policy of VLA models is eliminated.

Addressing the diversity of manipulation poses is crucial. While a fixed manipulation strategy can achieve high performance when the target's pose remains relatively constant, it fails to generalize across varied scenarios.
For optimal observation, position a camera fixed to the workbench facing the robot is recommended, as this setup minimizes self-occlusion compared to a head-mounted camera.

For action label used in model training, subtracting between current states rather than referencing control targets can yield greater benefits. The VLA model cannot observe the image at the control target, using control target will introduce another systematic error into the policy. Instead, it will automatically adjust the control error towards its desired pose over several steps.

\subsubsection{Stage Decoupling Necessity}
To assess the impact of the large-workspace \srp stage on model performance, we trained two models with and without \srp data, and then evaluated them with two different initial setups.
In the testing environments where the robot is initialized near the target, the end effector is randomly positioned within a maximum range of $0.15m$ from the target object. In the alternative setup, the robot starts in a random pose within the workspace. The results are shown in \cref{tab:approach_sr}. 

\begin{table}[tp]
  \centering
  \begin{tabular}{@{}cl|cc@{}}
    \toprule
    Train w/ \srp & Eval. Init. & Test Scenes & Zero-Shot \\
    \midrule
    \checkmark & Random & 0.53 & 0.40  \\
    \checkmark & Near Target & \textbf{0.80} & \textbf{0.68} \\
    $\times$ & Random & 0.08 & 0.06 \\
    $\times$ & Near Target & 0.73 & 0.66 \\
    \bottomrule
  \end{tabular}
  \caption{Influence of the \srp stage on the model performance.}
  \label{tab:approach_sr}
\end{table}

In both models, the evaluation revealed that the \srp stage significantly reduced performance. Although the \srp stage provides the necessary capability for the model to locate the target, its difficulty is much higher than that of the \pip stage. This increased difficulty stems from the vast exploration space, which introduces diverse situations. Consequently, it is advisable to learn the two stages in a decoupled manner, \ie train the two sub-tasks with different amount of data at varying speed, which is the the way our method employs.

\begin{figure}[tp]
  \centering
  \includegraphics[width=\linewidth, trim=2 2 2 2, clip]{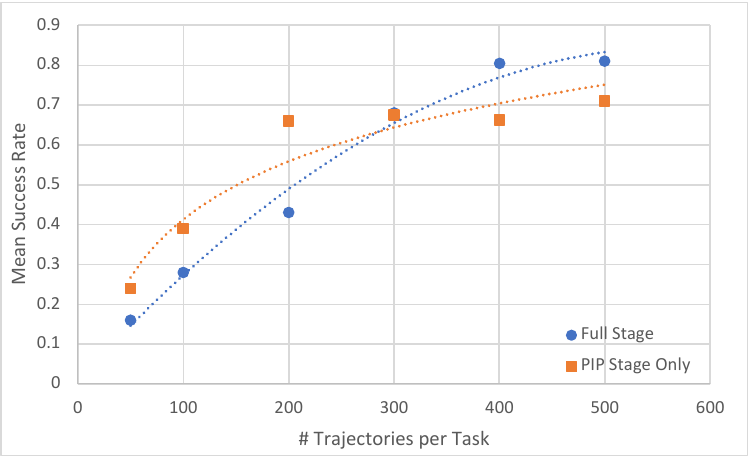}
  \caption{The scaling relationship between task performance and the data volume for models trained with and without the data of \srp segments.}
  \label{fig:scaling}
\end{figure}

To verify the impact of the wide-range workspace on task difficulty during the \srp stage, we trained two sets of models: one using full-stage trajectories and the other using only \pip segments. The results, as shown in \cref{fig:scaling}, indicate that the performance bottleneck, \ie the data volume required for the model to converge to a relatively stable success rate, occurs much earlier in the simpler manipulation-only tasks compared to the full-stage tasks. 
Together with the scaling law related to the target variety shown in \cref{fig:multi-task-result}, we observe that the difficulties of different sub-tasks arise from distinct aspects: the space-related \srp stage requires demonstrations that cover the entire workspace, whereas the geometry-related \pip stage necessitates a variety of target shapes to derive an effective manipulation strategy.
These findings highlight the necessity of preparing decoupled data for the various stages within a single task category.

\subsubsection{Data Balancing Strategies}
\label{sec:balancing}
During model training, we observed that repeating the full-stage trajectories can lead to a significant better performance. 
To verify which repeating method is best, we compared different ways of data repeating methods in \cref{tab:copy}. 
The results suggest that repeating the \pip segments of the full-stage data $\tau^{\pip} \subset \tau^{F}$ excluding the \srp segments, yields optimal results. This approach maintains distribution consistency among sub-tasks and implicitly regularizes the adaptation of sub-task weights, preserving the temporal dependencies between sub-tasks and stabilizing the model optimization process.

\begin{table}
  \centering
  \begin{tabular}{@{}l|cc@{}}
    \toprule
    & Test Scenes & Zero-Shot \\
    \midrule
    No Repeating & 0.58 & 0.46  \\
    $^\dag$Repeat $\tau^F$ & 0.59 & 0.41 \\
    $^\dag$Repeat $\tau^{\pip}$ & 0.59 & \textbf{0.58} \\
    \bottomrule
  \end{tabular}
  \caption{The performance of of models trained with different \pip segment repeating methods. All models use $p_{\srp} = 66\%$ of \pip segments during training. $^\dag$ The corresponding segments are duplicated $\left \lfloor \frac{N_2}{N_1} \right \rfloor = 2$ times.}
  \label{tab:copy}
\end{table}

\section{Conclusion}

In this paper we propose \pipelinename benchmark with the \projname method. The \pipelinename provides a cost-effective pipeline for generating and evaluating physically realistic manipulation trajectories. The \projname method is a stage-decoupled training paradigm that enhances VLA models through strategic utilization of additional cost-effective spatial reasoning data. Our key contributions are threefold: (1) Additional \srp data introduced to the decoupled costly full-stage trajectories acts as a catalytic role that can achieve a 41\% improvement in zero-shot success rate, by allowing enhanced training on spatial search patterns. This result, from another perspective, reduced the dependency on human-collected data. (2) The \srp/\pip mixture ratio follows a logarithmic law indicating at most 4x more additional \srp data can be added into the full-stage data to maximize the generalization performance. (3) The half-trajectories with only \srp data can serve as the bridge to transfer the generalized skills to novel target objects, providing a novel way in instance-wide generalization. 
 Our stage-decoupled paradigm opens new directions for stage-aware curriculum learning in embodied AI, particularly in adaptive stage boundary detection.


{
    \small
    \bibliographystyle{ieeenat_fullname}
    \bibliography{main}

\begin{thebibliography}{53}
\providecommand{\natexlab}[1]{#1}
\providecommand{\url}[1]{\texttt{#1}}
\expandafter\ifx\csname urlstyle\endcsname\relax
  \providecommand{\doi}[1]{doi: #1}\else
  \providecommand{\doi}{doi: \begingroup \urlstyle{rm}\Url}\fi

\bibitem[Black et~al.(2024)Black, Brown, Driess, Esmail, Equi, Finn, Fusai, Groom, Hausman, Ichter, et~al.]{black2410pi0}
Kevin Black, Noah Brown, Danny Driess, Adnan Esmail, Michael Equi, Chelsea Finn, Niccolo Fusai, Lachy Groom, Karol Hausman, Brian Ichter, et~al.
\newblock $\pi_0$: A vision-language-action flow model for general robot control.
\newblock \emph{arXiv preprint arXiv:2410.24164}, 2024.

\bibitem[Brohan et~al.(2022)Brohan, Brown, Carbajal, Chebotar, Dabis, Finn, Gopalakrishnan, Hausman, Herzog, Hsu, et~al.]{brohan2022rt1}
Anthony Brohan, Noah Brown, Justice Carbajal, Yevgen Chebotar, Joseph Dabis, Chelsea Finn, Keerthana Gopalakrishnan, Karol Hausman, Alex Herzog, Jasmine Hsu, et~al.
\newblock {RT-1: Robotics Transformer for Real-World Control at Scale}.
\newblock \emph{arXiv preprint arXiv:2212.06817}, 2022.

\bibitem[Brohan et~al.(2023)Brohan, Chebotar, Finn, Hausman, Herzog, Ho, Ibarz, Irpan, Jang, Julian, et~al.]{brohan2023can}
Anthony Brohan, Yevgen Chebotar, Chelsea Finn, Karol Hausman, Alexander Herzog, Daniel Ho, Julian Ibarz, Alex Irpan, Eric Jang, Ryan Julian, et~al.
\newblock {Do as I Can, Not as I Say: Grounding Language in Robotic Affordances}.
\newblock In \emph{Conference on robot learning}, pages 287--318. PMLR, 2023.

\bibitem[Cheang et~al.(2024)Cheang, Chen, Jing, Kong, Li, Li, Liu, Wu, Xu, Yang, et~al.]{cheang2024gr2}
Chi-Lam Cheang, Guangzeng Chen, Ya Jing, Tao Kong, Hang Li, Yifeng Li, Yuxiao Liu, Hongtao Wu, Jiafeng Xu, Yichu Yang, et~al.
\newblock {GR-2: A Generative Video-Language-Action Model with Web-Scale Knowledge for Robot Manipulation}.
\newblock \emph{arXiv preprint arXiv:2410.06158}, 2024.

\bibitem[Doshi et~al.(2024)Doshi, Walke, Mees, Dasari, and Levine]{doshi2024scaling}
Ria Doshi, Homer Walke, Oier Mees, Sudeep Dasari, and Sergey Levine.
\newblock {Scaling Cross-Embodied Learning: One Policy for Manipulation, Navigation, Locomotion and Aviation}.
\newblock \emph{arXiv preprint arXiv:2408.11812}, 2024.

\bibitem[Fang et~al.(2020)Fang, Wang, Gou, and Lu]{fang2020graspnet}
Hao-Shu Fang, Chenxi Wang, Minghao Gou, and Cewu Lu.
\newblock {GraspNet-1Billion: A Large-Scale Benchmark for General Object Grasping}.
\newblock In \emph{Proceedings of the IEEE/CVF conference on computer vision and pattern recognition}, pages 11444--11453, 2020.

\bibitem[Finn et~al.(2017)Finn, Abbeel, and Levine]{finn2017model}
Chelsea Finn, Pieter Abbeel, and Sergey Levine.
\newblock {Model-Agnostic Meta-Learning for Fast Adaptation of Deep Networks}.
\newblock In \emph{International conference on machine learning}, pages 1126--1135. PMLR, 2017.

\bibitem[Fu et~al.(2024)Fu, Zhao, and Finn]{fu2024mobile}
Zipeng Fu, Tony~Z Zhao, and Chelsea Finn.
\newblock {Mobile ALOHA: Learning Bimanual Mobile Manipulation With Low-Cost Whole-Body Teleoperation}.
\newblock \emph{arXiv preprint arXiv:2401.02117}, 2024.

\bibitem[Gong et~al.(2023)Gong, Huang, Zhao, Geng, Gao, Wu, Ai, Zhou, Terzopoulos, Zhu, et~al.]{gong2023arnold}
Ran Gong, Jiangyong Huang, Yizhou Zhao, Haoran Geng, Xiaofeng Gao, Qingyang Wu, Wensi Ai, Ziheng Zhou, Demetri Terzopoulos, Song-Chun Zhu, et~al.
\newblock {ARNOLD: A Benchmark for Language-Grounded Task Learning with Continuous States in Realistic 3D Scenes}.
\newblock In \emph{Proceedings of the IEEE/CVF International Conference on Computer Vision}, pages 20483--20495, 2023.

\bibitem[Ha et~al.(2023)Ha, Florence, and Song]{ha2023scaling}
Huy Ha, Pete Florence, and Shuran Song.
\newblock {Scaling Up and Distilling Down: Language-Guided Robot Skill Acquisition}.
\newblock In \emph{Conference on Robot Learning}, pages 3766--3777. PMLR, 2023.

\bibitem[Huang et~al.(2025)Huang, Chen, Zhou, Chen, Jiang, Hu, Gao, Li, Yao, and Ren]{huang2025enerverse}
Siyuan Huang, Liliang Chen, Pengfei Zhou, Shengcong Chen, Zhengkai Jiang, Yue Hu, Peng Gao, Hongsheng Li, Maoqing Yao, and Guanghui Ren.
\newblock {EnerVerse: Envisioning Embodied Future Space for Robotics Manipulation}.
\newblock \emph{arXiv preprint arXiv:2501.01895}, 2025.

\bibitem[James et~al.(2019)James, Wohlhart, Kalakrishnan, Kalashnikov, Irpan, Ibarz, Levine, Hadsell, and Bousmalis]{james2019sim}
Stephen James, Paul Wohlhart, Mrinal Kalakrishnan, Dmitry Kalashnikov, Alex Irpan, Julian Ibarz, Sergey Levine, Raia Hadsell, and Konstantinos Bousmalis.
\newblock {Sim-to-Real via Sim-to-Sim: Data-Efficient Robotic Grasping via Randomized-to-Canonical Adaptation Networks}.
\newblock In \emph{Proceedings of the IEEE/CVF conference on computer vision and pattern recognition}, pages 12627--12637, 2019.

\bibitem[Jin et~al.(2024)Jin, Li, Yong, Shi, Hao, Sun, Zhang, and Fang]{jin2024robotgpt}
Yixiang Jin, Dingzhe Li, A Yong, Jun Shi, Peng Hao, Fuchun Sun, Jianwei Zhang, and Bin Fang.
\newblock {RobotGPT: Robot Manipulation Learning from ChatGPT}.
\newblock \emph{IEEE Robotics and Automation Letters}, 9\penalty0 (3):\penalty0 2543--2550, 2024.

\bibitem[Ke et~al.(2024)Ke, Gkanatsios, and Fragkiadaki]{ke20243d}
Tsung-Wei Ke, Nikolaos Gkanatsios, and Katerina Fragkiadaki.
\newblock {3D Diffuser Actor: Policy Diffusion with 3D Scene Representations}.
\newblock \emph{arXiv preprint arXiv:2402.10885}, 2024.

\bibitem[Kim et~al.(2024)Kim, Pertsch, Karamcheti, Xiao, Balakrishna, Nair, Rafailov, Foster, Lam, Sanketi, et~al.]{kim2024openvla}
Moo~Jin Kim, Karl Pertsch, Siddharth Karamcheti, Ted Xiao, Ashwin Balakrishna, Suraj Nair, Rafael Rafailov, Ethan Foster, Grace Lam, Pannag Sanketi, et~al.
\newblock {OpenVLA: An Open-Source Vision-Language-Action Model}.
\newblock \emph{arXiv preprint arXiv:2406.09246}, 2024.

\bibitem[Laskin et~al.(2020)Laskin, Lee, Stooke, Pinto, Abbeel, and Srinivas]{laskin2020reinforcement}
Misha Laskin, Kimin Lee, Adam Stooke, Lerrel Pinto, Pieter Abbeel, and Aravind Srinivas.
\newblock {Reinforcement Learning with Augmented Data}.
\newblock \emph{Advances in neural information processing systems}, 33:\penalty0 19884--19895, 2020.

\bibitem[Lee et~al.(2024)Lee, Wang, Etukuru, Kim, Shafiullah, and Pinto]{lee2024behavior}
Seungjae Lee, Yibin Wang, Haritheja Etukuru, H~Jin Kim, Nur Muhammad~Mahi Shafiullah, and Lerrel Pinto.
\newblock {Behavior Generation with Latent Actions}.
\newblock \emph{arXiv preprint arXiv:2403.03181}, 2024.

\bibitem[Li et~al.(2023{\natexlab{a}})Li, Zhang, Wong, Gokmen, Srivastava, Mart{\'\i}n-Mart{\'\i}n, Wang, Levine, Lingelbach, Sun, et~al.]{li2023behavior}
Chengshu Li, Ruohan Zhang, Josiah Wong, Cem Gokmen, Sanjana Srivastava, Roberto Mart{\'\i}n-Mart{\'\i}n, Chen Wang, Gabrael Levine, Michael Lingelbach, Jiankai Sun, et~al.
\newblock {BEHAVIOR-1K: A Benchmark for Embodied AI with 1,000 Everyday Activities and Realistic Simulation}.
\newblock In \emph{Conference on Robot Learning}, pages 80--93. PMLR, 2023{\natexlab{a}}.

\bibitem[Li et~al.(2024{\natexlab{a}})Li, Liang, Wang, Luo, Chen, Liao, Wei, Deng, Xu, Zhang, et~al.]{li2024cogact}
Qixiu Li, Yaobo Liang, Zeyu Wang, Lin Luo, Xi Chen, Mozheng Liao, Fangyun Wei, Yu Deng, Sicheng Xu, Yizhong Zhang, et~al.
\newblock {CogACT: A Foundational Vision-Language-Action Model for Synergizing Cognition and Action in Robotic Manipulation}.
\newblock \emph{arXiv preprint arXiv:2411.19650}, 2024{\natexlab{a}}.

\bibitem[Li et~al.(2023{\natexlab{b}})Li, Liu, Zhang, Yu, Xu, Wu, Cheang, Jing, Zhang, Liu, et~al.]{li2023vision}
Xinghang Li, Minghuan Liu, Hanbo Zhang, Cunjun Yu, Jie Xu, Hongtao Wu, Chilam Cheang, Ya Jing, Weinan Zhang, Huaping Liu, et~al.
\newblock {Vision-Language Foundation Models as Effective Robot Imitators}.
\newblock \emph{arXiv preprint arXiv:2311.01378}, 2023{\natexlab{b}}.

\bibitem[Li et~al.(2024{\natexlab{b}})Li, Li, Liu, Wang, Liu, Kang, Ma, Kong, Zhang, and Liu]{li2024towards}
Xinghang Li, Peiyan Li, Minghuan Liu, Dong Wang, Jirong Liu, Bingyi Kang, Xiao Ma, Tao Kong, Hanbo Zhang, and Huaping Liu.
\newblock {Towards Generalist Robot Policies: What Matters in Building Vision-Language-Action Models}.
\newblock \emph{arXiv preprint arXiv:2412.14058}, 2024{\natexlab{b}}.

\bibitem[Lin et~al.(2024)Lin, Hu, Sheng, Wen, You, and Gao]{lin2024data}
Fanqi Lin, Yingdong Hu, Pingyue Sheng, Chuan Wen, Jiacheng You, and Yang Gao.
\newblock {Data Scaling Laws in Imitation Learning for Robotic Manipulation}.
\newblock \emph{arXiv preprint arXiv:2410.18647}, 2024.

\bibitem[Liu et~al.(2024{\natexlab{a}})Liu, Yan, Zheng, Feng, Huang, and Ma]{liu2024robouniview}
Fanfan Liu, Feng Yan, Liming Zheng, Chengjian Feng, Yiyang Huang, and Lin Ma.
\newblock {RoboUniView: Visual-Language Model with Unified View Representation for Robotic Manipulation}.
\newblock \emph{arXiv preprint arXiv:2406.18977}, 2024{\natexlab{a}}.

\bibitem[Liu et~al.(2024{\natexlab{b}})Liu, Liu, Wang, Lee, Zhou, An, Yang, Zhang, Guo, and Zhang]{liu2024robomamba}
Jiaming Liu, Mengzhen Liu, Zhenyu Wang, Lily Lee, Kaichen Zhou, Pengju An, Senqiao Yang, Renrui Zhang, Yandong Guo, and Shanghang Zhang.
\newblock {RoboMamba: Multimodal State Space Model for Efficient Robot Reasoning and Manipulation}.
\newblock \emph{arXiv preprint arXiv:2406.04339}, 2024{\natexlab{b}}.

\bibitem[Liu et~al.(2024{\natexlab{c}})Liu, Wu, Li, Tan, Chen, Wang, Xu, Su, and Zhu]{liu2024rdt}
Songming Liu, Lingxuan Wu, Bangguo Li, Hengkai Tan, Huayu Chen, Zhengyi Wang, Ke Xu, Hang Su, and Jun Zhu.
\newblock {RDT-1B: A Diffusion Foundation Model for Bimanual Manipulation}.
\newblock \emph{arXiv preprint arXiv:2410.07864}, 2024{\natexlab{c}}.

\bibitem[Mandlekar et~al.(2023)Mandlekar, Nasiriany, Wen, Akinola, Narang, Fan, Zhu, and Fox]{mandlekar2023mimicgen}
Ajay Mandlekar, Soroush Nasiriany, Bowen Wen, Iretiayo Akinola, Yashraj Narang, Linxi Fan, Yuke Zhu, and Dieter Fox.
\newblock {MimicGen: A Data Generation System for Scalable Robot Learning Using Human Demonstrations}.
\newblock \emph{arXiv preprint arXiv:2310.17596}, 2023.

\bibitem[Mees et~al.(2022)Mees, Hermann, Rosete-Beas, and Burgard]{mees2022calvin}
Oier Mees, Lukas Hermann, Erick Rosete-Beas, and Wolfram Burgard.
\newblock {CALVIN: A Benchmark for Language-Conditioned Policy Learning for Long-Horizon Robot Manipulation Tasks}.
\newblock \emph{IEEE Robotics and Automation Letters}, 7\penalty0 (3):\penalty0 7327--7334, 2022.

\bibitem[Mees et~al.(2023)Mees, Borja-Diaz, and Burgard]{mees2023hulc2}
Oier Mees, Jessica Borja-Diaz, and Wolfram Burgard.
\newblock {Grounding Language with Visual Affordances Over Unstructured Data}.
\newblock In \emph{2023 IEEE International Conference on Robotics and Automation (ICRA)}, pages 11576--11582. IEEE, 2023.

\bibitem[Mete et~al.(2025)Mete, Xue, Wilcox, Chen, and Garg]{mete2025quest}
Atharva Mete, Haotian Xue, Albert Wilcox, Yongxin Chen, and Animesh Garg.
\newblock {QueST: Self-Supervised Skill Abstractions for Learning Continuous Control}.
\newblock \emph{Advances in Neural Information Processing Systems}, 37:\penalty0 4062--4089, 2025.

\bibitem[Mittal et~al.(2023)Mittal, Yu, Yu, Liu, Rudin, Hoeller, Yuan, Singh, Guo, Mazhar, Mandlekar, Babich, State, Hutter, and Garg]{mittal2023orbit}
Mayank Mittal, Calvin Yu, Qinxi Yu, Jingzhou Liu, Nikita Rudin, David Hoeller, Jia~Lin Yuan, Ritvik Singh, Yunrong Guo, Hammad Mazhar, Ajay Mandlekar, Buck Babich, Gavriel State, Marco Hutter, and Animesh Garg.
\newblock {Orbit: A Unified Simulation Framework for Interactive Robot Learning Environments}.
\newblock \emph{IEEE Robotics and Automation Letters}, 8\penalty0 (6):\penalty0 3740--3747, 2023.

\bibitem[Mo et~al.(2019)Mo, Zhu, Chang, Yi, Tripathi, Guibas, and Su]{Mo_2019_CVPR}
Kaichun Mo, Shilin Zhu, Angel~X. Chang, Li Yi, Subarna Tripathi, Leonidas~J. Guibas, and Hao Su.
\newblock {PartNet: A Large-Scale Benchmark for Fine-Grained and Hierarchical Part-Level 3D Object Understanding}.
\newblock In \emph{Proceedings of the IEEE Conference on Computer Vision and Pattern Recognition (CVPR)}, 2019.

\bibitem[Nair et~al.(2022)Nair, Rajeswaran, Kumar, Finn, and Gupta]{nair2022r3m}
Suraj Nair, Aravind Rajeswaran, Vikash Kumar, Chelsea Finn, and Abhinav Gupta.
\newblock {R3M: A Universal Visual Representation for Robot Manipulation}.
\newblock \emph{arXiv preprint arXiv:2203.12601}, 2022.

\bibitem[Nguyen(1986)]{Nguyen1986}
V.-D. Nguyen.
\newblock {Constructing Force-Closure Grasps}.
\newblock In \emph{Proceedings of the IEEE International Conference on Robotics and Automation}, pages 1368--1373, 1986.

\bibitem[NVIDIA(2023{\natexlab{a}})]{isaacsim}
NVIDIA.
\newblock {Isaac Sim - Robotics Simulation and Synthetic Data Generation}.
\newblock \url{https://developer.nvidia.com/isaac/sim}, 2023{\natexlab{a}}.
\newblock Dec 16, 2024.

\bibitem[NVIDIA(2023{\natexlab{b}})]{nucleus}
NVIDIA.
\newblock {Omniverse Nucleus}.
\newblock \url{https://docs.omniverse.nvidia.com/nucleus/latest/index.html}, 2023{\natexlab{b}}.
\newblock Dec 16, 2024.

\bibitem[O’Neill et~al.(2024)O’Neill, Rehman, Maddukuri, Gupta, Padalkar, Lee, Pooley, Gupta, Mandlekar, Jain, et~al.]{rt-x}
Abby O’Neill, Abdul Rehman, Abhiram Maddukuri, Abhishek Gupta, Abhishek Padalkar, Abraham Lee, Acorn Pooley, Agrim Gupta, Ajay Mandlekar, Ajinkya Jain, et~al.
\newblock {Open X-Embodiment: Robotic Learning Datasets and RT-X Models}.
\newblock In \emph{2024 IEEE International Conference on Robotics and Automation (ICRA)}, pages 6892--6903. IEEE, 2024.

\bibitem[Reuss et~al.(2024)Reuss, Ya{\u{g}}murlu, Wenzel, and Lioutikov]{reuss2024multimodal}
Moritz Reuss, {\"O}mer~Erdin{\c{c}} Ya{\u{g}}murlu, Fabian Wenzel, and Rudolf Lioutikov.
\newblock {Multimodal Diffusion Transformer: Learning Versatile Behavior from Multimodal Goals}.
\newblock \emph{arXiv preprint arXiv:2407.05996}, 2024.

\bibitem[Sermanet et~al.(2024)Sermanet, Ding, Zhao, Xia, Dwibedi, Gopalakrishnan, Chan, Dulac-Arnold, Maddineni, Joshi, et~al.]{sermanet2024robovqa}
Pierre Sermanet, Tianli Ding, Jeffrey Zhao, Fei Xia, Debidatta Dwibedi, Keerthana Gopalakrishnan, Christine Chan, Gabriel Dulac-Arnold, Sharath Maddineni, Nikhil~J Joshi, et~al.
\newblock {RoboVQA: Multimodal Long-Horizon Reasoning for Robotics}.
\newblock In \emph{2024 IEEE International Conference on Robotics and Automation (ICRA)}, pages 645--652. IEEE, 2024.

\bibitem[Sundaralingam et~al.(2023)Sundaralingam, Hari, Fishman, Garrett, Van~Wyk, Blukis, Millane, Oleynikova, Handa, Ramos, Ratliff, and Fox]{curobo_icra23}
Balakumar Sundaralingam, Siva Kumar~Sastry Hari, Adam Fishman, Caelan Garrett, Karl Van~Wyk, Valts Blukis, Alexander Millane, Helen Oleynikova, Ankur Handa, Fabio Ramos, Nathan Ratliff, and Dieter Fox.
\newblock {CuRobo: Parallelized Collision-Free Robot Motion Generation}.
\newblock In \emph{2023 IEEE International Conference on Robotics and Automation (ICRA)}, pages 8112--8119, 2023.

\bibitem[Szot et~al.(2025)Szot, Mazoure, Agrawal, Hjelm, Kira, and Toshev]{szot2025grounding}
Andrew Szot, Bogdan Mazoure, Harsh Agrawal, R~Devon Hjelm, Zsolt Kira, and Alexander Toshev.
\newblock {Grounding Multimodal Large Language Models in Actions}.
\newblock \emph{Advances in Neural Information Processing Systems}, 37:\penalty0 20198--20224, 2025.

\bibitem[Tan et~al.(2024)Tan, Xu, Ying, Mao, Liu, Zhang, Su, and Zhu]{tan2024manibox}
Hengkai Tan, Xuezhou Xu, Chengyang Ying, Xinyi Mao, Songming Liu, Xingxing Zhang, Hang Su, and Jun Zhu.
\newblock {ManiBox: Enhancing Spatial Grasping Generalization via Scalable Simulation Data Generation}.
\newblock \emph{arXiv preprint arXiv:2411.01850}, 2024.

\bibitem[Tao et~al.(2024)Tao, Xiang, Shukla, Qin, Hinrichsen, Yuan, Bao, Lin, Liu, kai Chan, Gao, Li, Mu, Xiao, Gurha, Huang, Calandra, Chen, Luo, and Su]{taomaniskill3}
Stone Tao, Fanbo Xiang, Arth Shukla, Yuzhe Qin, Xander Hinrichsen, Xiaodi Yuan, Chen Bao, Xinsong Lin, Yulin Liu, Tse kai Chan, Yuan Gao, Xuanlin Li, Tongzhou Mu, Nan Xiao, Arnav Gurha, Zhiao Huang, Roberto Calandra, Rui Chen, Shan Luo, and Hao Su.
\newblock {ManiSkill3: GPU Parallelized Robotics Simulation and Rendering for Generalizable Embodied AI}.
\newblock \emph{arXiv preprint arXiv:2410.00425}, 2024.

\bibitem[Walke et~al.(2023)Walke, Black, Zhao, Vuong, Zheng, Hansen-Estruch, He, Myers, Kim, Du, et~al.]{walke2023bridgedata}
Homer~Rich Walke, Kevin Black, Tony~Z Zhao, Quan Vuong, Chongyi Zheng, Philippe Hansen-Estruch, Andre~Wang He, Vivek Myers, Moo~Jin Kim, Max Du, et~al.
\newblock {BridgeData V2: A Dataset for Robot Learning at Scale}.
\newblock In \emph{Conference on Robot Learning}, pages 1723--1736. PMLR, 2023.

\bibitem[Wang et~al.(2025)Wang, Chen, Zhao, and He]{wang2025scaling}
Lirui Wang, Xinlei Chen, Jialiang Zhao, and Kaiming He.
\newblock {Scaling Proprioceptive-Visual Learning with Heterogeneous Pre-Trained Transformers}.
\newblock \emph{Advances in Neural Information Processing Systems}, 37:\penalty0 124420--124450, 2025.

\bibitem[Wang et~al.(2023)Wang, Xian, Chen, Wang, Wang, Fragkiadaki, Erickson, Held, and Gan]{wang2023robogen}
Yufei Wang, Zhou Xian, Feng Chen, Tsun-Hsuan Wang, Yian Wang, Katerina Fragkiadaki, Zackory Erickson, David Held, and Chuang Gan.
\newblock {RoboGen: Towards Unleashing Infinite Data for Automated Robot Learning via Generative Simulation}.
\newblock \emph{arXiv preprint arXiv:2311.01455}, 2023.

\bibitem[Wu et~al.(2024)Wu, Hou, Liu, Che, Ju, Yang, Li, Zhao, Xu, Yang, et~al.]{wu2024robomind}
Kun Wu, Chengkai Hou, Jiaming Liu, Zhengping Che, Xiaozhu Ju, Zhuqin Yang, Meng Li, Yinuo Zhao, Zhiyuan Xu, Guang Yang, et~al.
\newblock {RoboMIND: Benchmark on Multi-Embodiment Intelligence Normative Data for Robot Manipulation}.
\newblock \emph{arXiv preprint arXiv:2412.13877}, 2024.

\bibitem[Yan et~al.(2024)Yan, Liu, Zheng, Zhong, Huang, Guan, Feng, and Ma]{yan2024robomm}
Feng Yan, Fanfan Liu, Liming Zheng, Yufeng Zhong, Yiyang Huang, Zechao Guan, Chengjian Feng, and Lin Ma.
\newblock {RoboMM: All-in-One Multimodal Large Model for Robotic Manipulation}.
\newblock \emph{arXiv preprint arXiv:2412.07215}, 2024.

\bibitem[Zhao et~al.(2023)Zhao, Kumar, Levine, and Finn]{zhao2023act}
Tony~Z Zhao, Vikash Kumar, Sergey Levine, and Chelsea Finn.
\newblock {Learning Fine-Grained Bimanual Manipulation with Low-Cost Hardware}.
\newblock \emph{arXiv preprint arXiv:2304.13705}, 2023.

\bibitem[Zhen et~al.(2024)Zhen, Qiu, Chen, Yang, Yan, Du, Hong, and Gan]{zhen20243d}
Haoyu Zhen, Xiaowen Qiu, Peihao Chen, Jincheng Yang, Xin Yan, Yilun Du, Yining Hong, and Chuang Gan.
\newblock {3D-VLA: A 3D Vision-Language-Action Generative World Model}.
\newblock \emph{arXiv preprint arXiv:2403.09631}, 2024.

\bibitem[Zheng et~al.(2024)Zheng, Yan, Liu, Feng, Kang, and Ma]{zheng2024robocas}
Liming Zheng, Feng Yan, Fanfan Liu, Chengjian Feng, Zhuoliang Kang, and Lin Ma.
\newblock {RoboCAS: A Benchmark for Robotic Manipulation in Complex Object Arrangement Scenarios}.
\newblock \emph{arXiv preprint arXiv:2407.06951}, 2024.

\bibitem[Zhu et~al.(2024)Zhu, Yang, Wang, Yang, Wang, and He]{zhu2024spa}
Haoyi Zhu, Honghui Yang, Yating Wang, Jiange Yang, Limin Wang, and Tong He.
\newblock {SPA: 3D Spatial-Awareness Enables Effective Embodied Representation}.
\newblock \emph{arXiv preprint arXiv:2410.08208}, 2024.

\bibitem[Zhu et~al.(2025)Zhu, Zhu, Li, Zhou, Wen, Liu, Shen, Peng, and Feng]{zhu2025objectvla}
Minjie Zhu, Yichen Zhu, Jinming Li, Zhongyi Zhou, Junjie Wen, Xiaoyu Liu, Chaomin Shen, Yaxin Peng, and Feifei Feng.
\newblock {ObjectVLA: End-to-End Open-World Object Manipulation Without Demonstration}.
\newblock \emph{arXiv preprint arXiv:2502.19250}, 2025.

\bibitem[Zitkovich et~al.(2023)Zitkovich, Yu, Xu, Xu, Xiao, Xia, Wu, Wohlhart, Welker, Wahid, et~al.]{zitkovich2023rt2}
Brianna Zitkovich, Tianhe Yu, Sichun Xu, Peng Xu, Ted Xiao, Fei Xia, Jialin Wu, Paul Wohlhart, Stefan Welker, Ayzaan Wahid, et~al.
\newblock {RT-2: Vision-Language-Action Models Transfer Web Knowledge to Robotic Control}.
\newblock In \emph{Conference on Robot Learning}, pages 2165--2183. PMLR, 2023.

\end{thebibliography}
}

\clearpage
\appendix
\section{\pipelinename Benchmark Details}
\label{sec:data-benchmark}

Our environment is built on the NVIDIA Isaac Sim~\cite{isaacsim} simulator with the IsaacLab~\cite{mittal2023orbit} framework. Currently, parallel simulation is not supported in our benchmark. All images are rendered through the ray tracing pipeline.
Our pipeline supports manipulation trajectories generation on single-armed robots with both mobile or fixed bases, and also provides a framework to add novel target objects and novel tasks in the generation task. However, we only use data collected on robots with fixed base in our experiments.

\subsection{Environment Setup}

\paragraph{Simulator} Our \projname benchmark is built on NVIDIA Isaac Sim~\cite{isaacsim} platform, owing to its powerful physics simulation and image rendering capabilities. Photo-realistic observation on commonplace objects with diverse geometries can be effectively generated together with physically-accurate manipulation trajectory on this platform.

\paragraph{Scene} Our environment contains multiple workbenches in well-appointed rooms, allowing the agent to navigate between the table and manipulate the randomly scattered target objects on those tables. The texture of the room and tables can be randomly changed for data augmentation.

\paragraph{Agent} An agent with a 7-DoF Franka Emika Panda arm mounted on a fixed or mobile base is initialized on a proper position with a random arm pose. By default we put five RGB-D cameras in the environment for comprehensive observation, including
\begin{itemize}
    \item Wrist camera, mounted on the end effector of the agent;
    \item Base camera, mounted on the left shoulder position of the robot base to avoid the occlusion of the arm; 
    \item Side camera, mounted on the left front side of the agent and staring at the space in front of the arm which is the main workspace of the agent; 
    \item Static camera, mounted on a wall of the room providing a global view of the whole environment; 
    \item Table camera, placed on each workbench monitoring the status on this workspace.
\end{itemize}
Novel robots, together with cameras on other viewpoints, can also be easily applied in the environment just by editing few parameters in the configuration.


\paragraph{Manipulatable Objects} We provide 165 manipulatable objects in 89 types in our benchmark that are frequently encountered in everyday situations (see \cref{sec:assets} for details), and each object $\mathcal{M}$ are annotated with a large quantity of grasp pose labels $\mathcal{G}_{\mathcal{M}}$ through the method introduced in \cref{sec:label-annotation} to generate flexible manipulation trajectories that can meet the constraints of local scene and robot kinematics.

\subsection{Trajectory Generation}
\label{sec:traj-gen}

The data generation pipeline is decomposed into a two-level hierarchy: object-centric planning and agent-centric planning. We integrate the task steps using finite state machines to generate long-horizon manipulation data on objects with totally diverse geometries. Furthermore, this architecture enhances the scalability across various task types and supports the future expansion of automatic task construction. Generally, then trajectory generation process contains the following steps:

\paragraph{Scene Randomization} Given a scene pre-defined scene configuration template, before each generation process our benchmark randomly samples several manipulatable objects to be loaded in the environment at any position and orientation on the workbenches, among which one of them is selected as the target object $\mathcal{M}_{\text{tgt}}$. The agent is initialized at a random arm pose, and a random position in the environment for mobile agents. Optionally texture of the room can also be randomized.

\paragraph{Object-Centric Planning} Our benchmark uses key frames to guide trajectory generation, as a result we sample key poses of the agent, \eg grasp pose, drop pose, navigation site, through object centric planning. For manipulating step that requires high accuracy, pose labels $\mathcal{G}_{\mathcal{M}}$ are projected into the current scene using the pose of $\mathcal{M}_{\text{tgt}}$, and subsequently collision detection of gripper model at each $\mathbf{g} \in \mathcal{G}_{\mathcal{M}}$ with the whole scene is performed through the pipeline provided by CuRobo~\cite{curobo_icra23}. The manipulate pose is then sampled from collision-free labels. Pose offsets like pre-grasp offset, which we found crucial for BC, are added as external key frames in our pipeline.

\paragraph{Agent-Centric Planning} We employ CuRobo~\cite{curobo_icra23} for collision-free path planning across the key frame poses, and the path point is generated at the frequency of $120Hz$. Additional constraints can also be applied in the planing process for the reasonableness of trajectories. For 3-DoF trajectory of mobile base, the CuRobo benchmark is further modified to satisfy its inherent constraint.

\paragraph{Trajectory Collection} The interaction between objects in the scene and the agent are completely achieved through physics simulation. During each trajectory, language description of the whole task and and current subtask, together with other information for model training and trajectory replay, are recorded for direct use or further process. By default trajectory data is recorded at $30Hz$. 




\subsection{Assets}
\label{sec:assets}

We acquire the raw object model assets from GraspNet-1Billion~\cite{fang2020graspnet}, ARNOLD~\cite{gong2023arnold}, PartNet-Mobility~\cite{Mo_2019_CVPR} and Omniverse Nucleus~\cite{nucleus}. The texture assets are from ARNOLD~\cite{gong2023arnold} and Omniverse Nucleus~\cite{nucleus}.

To meet the requirements of physical simulation of Isaac Sim, we transfer the raw assets into the Universal Scene Description (USD) format and applied some properties on the model. For manipulatable objects selected from the models, we annotated diverse grasp labels (see \cref{sec:label-annotation}), object name and size information on the model for manipulation trajectory generation, and the overview of objects are shown in \cref{tab:obj-list}.

\begin{table*}[tp]
    \caption{List of manipulatable object types}
    \label{tab:obj-list}
    \centering
    \begin{tabular}{cllc|cllc}
        \toprule
        & Type & Obj. Name & Count & & Type & Obj. Name & Count \\ 
        \midrule
        1 & \multirow{10}{*}{Boxes} & Biscuit & 1 & 46 & \multirow{3}{*}{Tableware} & Bowl & 2 \\
        2 & & Sugar & 1 & 47 & & Mug & 3 \\
        3 & & Soap Box & 1 & 48 & & Knife & 3 \\ 
        \cmidrule{5-8} 
        4 & & Toothpaste Box & 1 & 49 & \multirow{26}{*}{Tools} & Component & 6 \\
        5 & & Milk & 1 & 50 & & Drill & 1 \\
        6 & & Battery Case & 1 & 51 & & Scissors & 2 \\
        7 & & Cardboard Cube & 1 & 52 & & Screwdriver & 2 \\
        8 & & Mark Pen Set & 1 & 53 & & Sharp Drill & 1 \\
        9 & & Block & 6 & 54 & & Screw & 3 \\
        10 & & Shoe Box & 1 & 55 & & Lock & 1 \\ 
        \cmidrule{1-4}
        11 & \multirow{5}{*}{Cans} & Soup Can & 1 & 56 & & Mouse & 2 \\
        12 & & Launch Meat & 1 & 57 & & Tape & 4 \\
        13 & & Potato Chip Can & 1 & 58 & & Ice Box & 1 \\
        14 & & Paint Can & 1 & 59 & & Art Brush & 1 \\
        15 & & Thin Can & 2 & 60 & & Battery & 2 \\ 
        \cmidrule{1-4}
        16 & \multirow{13}{*}{Bottles} & Toner & 1 & 61 & & Paint Brush Bridge & 1 \\
        17 & & Shampoo Tube & 5 & 62 & & Pen & 4 \\
        18 & & Facial Soup Bottle & 3 & 63 & & Sharpener & 2 \\
        19 & & Laundry Detergent & 2 & 64 & & Clock & 1 \\
        20 & & Toothpaste Tube & 1 & 65 & & Chalk & 7 \\
        21 & & Bottle with Pump & 5 & 66 & & Funnel & 1 \\
        22 & & Milk Tea & 1 & 67 & & Ruler Slide & 1 \\
        23 & & Shampoo Bottle & 2 & 68 & & Cutter & 1 \\
        24 & & Toothpaste Can & 1 & 69 & & Bucket & 9 \\
        25 & & Glass Jar & 3 & 70 & & Paint Tray & 1 \\
        26 & & Vase & 1 & 71 & & Glasses & 1 \\
        27 & & Paint & 4 & 72 & & Tongs & 1 \\
        28 & & Phial & 3 & 73 & & Chair & 1 \\ 
        \cmidrule{1-4}
        29 & \multirow{17}{*}{Toys} & Blue Ball & 1 & 74 & & Cap & 3 \\ 
        \cmidrule{5-8} 
        30 & & Blue Plasticine & 1 & 75 & \multirow{3}{*}{Boards} & Board & 3 \\
        31 & & Yellow Plasticine & 1 & 76 & & Small Board & 1 \\
        32 & & Torpedo & 1 & 77 & & Paper & 2 \\ 
        \cmidrule{5-8} 
        33 & & Water Gun & 1 & 78 & \multirow{4}{*}{Books} & Closed Book & 3 \\
        34 & & Dragon & 1 & 79 & & Opened Book & 1 \\
        35 & & Camel & 1 & 80 & & Book Pile & 2 \\
        36 & & Moose & 1 & 81 & & Notes & 5 \\ 
        \cmidrule{5-8} 
        37 & & Zebra & 1 & 82 & \multirow{8}{*}{Fruits} & Banana & 1 \\
        38 & & Elephant & 2 & 83 & & Strawberry & 1 \\
        39 & & Rhino & 1 & 84 & & Apple & 1 \\
        40 & & Chimpanzee & 2 & 85 & & Lemon & 1 \\
        41 & & Giraffe & 1 & 86 & & Peach & 1 \\
        42 & & Lion & 1 & 87 & & Pear & 1 \\
        43 & & Hippo & 1 & 88 & & Orange & 1 \\
        44 & & Magic Cube & 1 & 89 & & Plum & 1 \\
        45 & & Fan & 1 & & & & \\ 
        \bottomrule
    \end{tabular}
\end{table*}

\subsection{Grasp Label Annotation}
\label{sec:label-annotation}

Given the USD model of the manipulatable objects, we annotated an enormous amount of force-closure~\cite{Nguyen1986} grasp labels on each model similar to \cite{fang2020graspnet}, enabling our benchmark to generate diverse physically feasible manipulation poses on the object, which can actually be used in real-world operations.

\begin{figure}[tp]
    \centering
    \includegraphics[width=\linewidth, trim=250 60 250 60, clip]{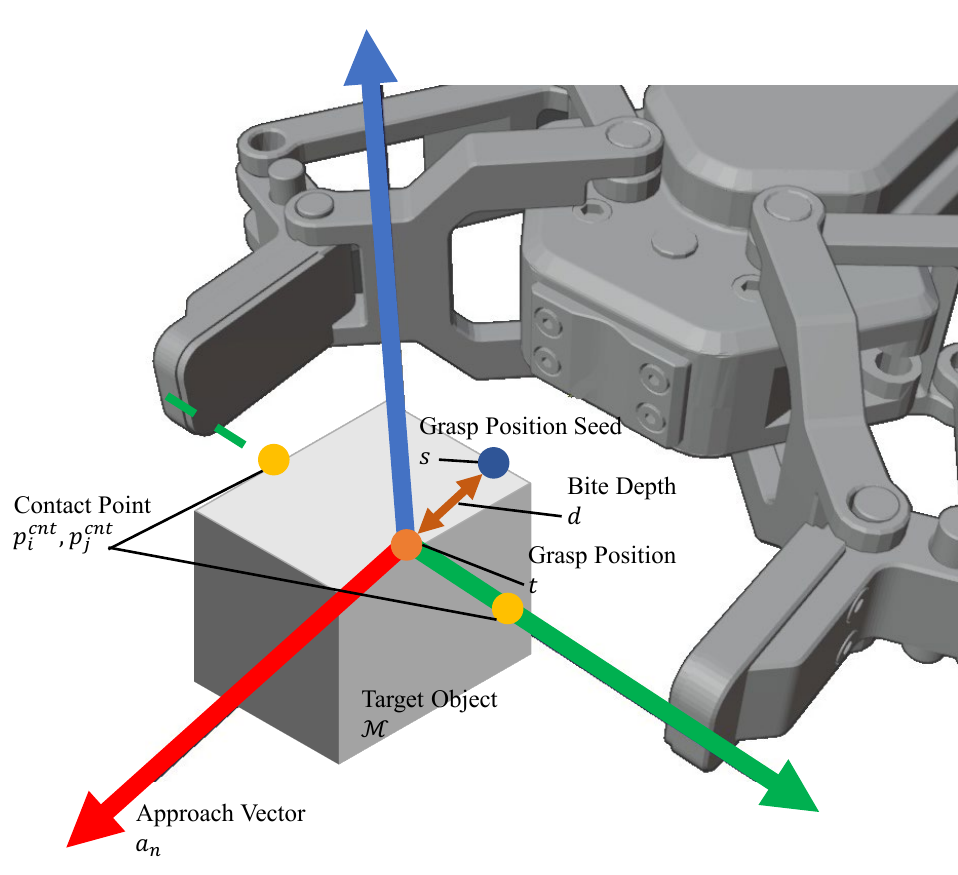}
    \caption{Grasp annotations}
    \label{fig:grasp-annotation}
\end{figure}

In order to generate initial grasp pose candidates $\mathcal{G}^{\text{cand}}_{\mathcal{M}} = \left\{ \mathbf{g}^{\text{cand}} = \left( \mathbf{R}, \mathbf{t} \right) \in SE(3) \right\}$, as shown in \cref{fig:grasp-annotation}, we first need to sample possible rotations $\mathbf{R}$ of $\mathbf{g}^{\text{cand}}$. Using Fibonacci lattice $N$ grasp approach vectors $\mathbf{a}_{n} = \left( x_{n}, y_{n}, z_{n} \right)^{T}$ can be sampled:
\begin{equation}
    \begin{aligned}
        z_{n} &= \frac{2n-1}{N} - 1, \\
        x_{n} &= \sqrt{1 - z_{n}^{2}} \cos{2 \pi n \phi}, \\
        y_{n} &= \sqrt{1 - z_{n}^{2}} \sin{2 \pi n \phi},
    \end{aligned}
\end{equation}
where $\phi = \frac{\sqrt{5} - 1}{2}$ is a constant. Subsequently, we evenly sample $K$ rotation angles in a semicircle and rotate the grasp pose around each $\mathbf{a}_{n}$ by this angle, in this way we totally get $N \times K$ grasp rotations $\mathbf{R}$. Given an object model $\mathcal{M}$, we sample $M$ grasp position seeds $\mathbf{s}$ around the surface of $\mathcal{M}$, and at each seed we sample $D$ offset distance alongside each $\mathbf{a}_{n}$ to obtain grasp positions with different bite depth, \ie, grasp position $\mathbf{t} = \mathbf{s} + \mathbf{a}_{n} d$. Thus, we totally generate $D \times M \times N \times K$ grasp poses in $\mathcal{G}^{\text{cand}}_{\mathcal{M}}$ for model $\mathcal{M}$.

To check whether a grasp candidate $\mathbf{g}^{\text{cand}} \in \mathcal{G}^{\text{cand}}_{\mathcal{M}}$ can successfully grasp the object $\mathcal{M}$ without collision, we first generate the surface point cloud $\mathbf{P}_{\mathcal{M}}=\left\{ \left( \mathbf{p}_{i}, \mathbf{n}_{i} \right) \right\}$ using Isaac Sim Replicator, where $\mathbf{p}_{i} \in \mathbb{R}^{3}$ is the position of the point under the model frame and $\mathbf{n}_{i} \in \mathbb{R}^{3}$ is the outward surface normal at $\mathbf{p}_{i}$. In order to get collision-free grasp poses, we simplify the two-finger gripper into three cubes: two fingers and a gripper hand (marked green in \cref{fig:grasp-annotation}), ensuring that there are not point $\mathbf{p}_{i}$ in the cubes, and at the same time the bite depth $d_{bite}$ is greater than a threshold $d_{min}$. Further more, to check whether the collision-free grasp pose $\mathbf{g}^{\text{cand}}$ is force-closure, we select the points closest to the two fingers $\mathbf{p}_{i}^{\text{cnt}}$, $\mathbf{p}_{j}^{\text{cnt}}$ as contact points, and check the angle between their surface normal $\mathbf{n}_{i}^{\text{cnt}}$, $\mathbf{n}_{j}^{\text{cnt}}$ and the vector $\mathbf{p}_{i}^{\text{cnt}} - \mathbf{p}_{j}^{\text{cnt}}$, \ie, $\mathbf{g}^{\text{cand}}$ is valid if
\begin{equation}
    \begin{cases}
        \tan \left \langle \mathbf{n}_{i}^{\text{cnt}}, \mathbf{p}_{i}^{\text{cnt}} - \mathbf{p}_{j}^{\text{cnt}} \right \rangle < \mu, \\
        \tan \left \langle \mathbf{n}_{j}^{\text{cnt}}, \mathbf{p}_{j}^{\text{cnt}} - \mathbf{p}_{i}^{\text{cnt}} \right \rangle < \mu,
    \end{cases}
\end{equation}
where $\mu$ is the static friction coefficient of the model surface, and $\left \langle \cdot, \cdot \right \rangle$ calculates the angle between the two vectors. In practice we set $\mu = 0.2$ to filter out labels with the highest quality. After such filtering process, the final grasp label annotation $\mathcal{G}_{\mathcal{M}} \subseteq \mathcal{G}_{\mathcal{M}}^{\text{cand}}$ that can successfully grasp the model $\mathcal{M}$ in physically world can be obtained, which can be used in the generation of diverse manipulation trajectories of $\mathcal{M}$.

\subsection{Additional Parameters in Dataset Generation}
\label{sec:data-parameters}
The range of random manipulation pose offset $\Delta \mathbf{T} = \left(\Delta\mathbf{R}, \Delta \mathbf{t}\right)$ in \cref{sec:srp-data-collection} is defined in Euler angles and translations, \wrt the Cartesian frame of the grasp pose $\mathbf{g}$ shown in \cref{fig:grasp-annotation}, \ie
\begin{equation}
    \begin{aligned}
        \Delta \mathbf{R} &= \text{rpy2mat} \left(\Delta roll, \Delta pitch, \Delta yaw \right), \\
        \Delta \mathbf{t} &= \left(\Delta x, \Delta y, \Delta z\right).
    \end{aligned}
\end{equation}
The range of parameters are shown in \cref{tab:data-params}.
\begin{table}[t]
    \centering
    \begin{tabular}{c|c}
        \toprule
        Parameter & Value \\
        \midrule
        $\Delta x$ ($m$) & $[0, 0.15]$ \\
        $\Delta y$ ($m$) & $[-0.05, 0.05]$ \\
        $\Delta z$ ($m$) & $[-0.05, 0.05]$ \\
        $\Delta roll$ ($rad$) & $[-\frac{\pi}{4}, \frac{\pi}{4}]$ \\
        $\Delta pitch$ ($rad$) & $[-\frac{\pi}{8}, \frac{\pi}{8}]$ \\
        $\Delta yaw$ ($rad$) & $[-\frac{\pi}{8}, \frac{\pi}{8}]$ \\
        \bottomrule
    \end{tabular}
    \caption{Parameters used in our data collection pipeline.}
    \label{tab:data-params}
\end{table}
\section{Training Data}

\subsection{Our Dataset}
\label{sec:collection}

\begin{figure}[tp]
    \centering
    \includegraphics[width=0.9\linewidth, trim=170 0 170 0, clip]{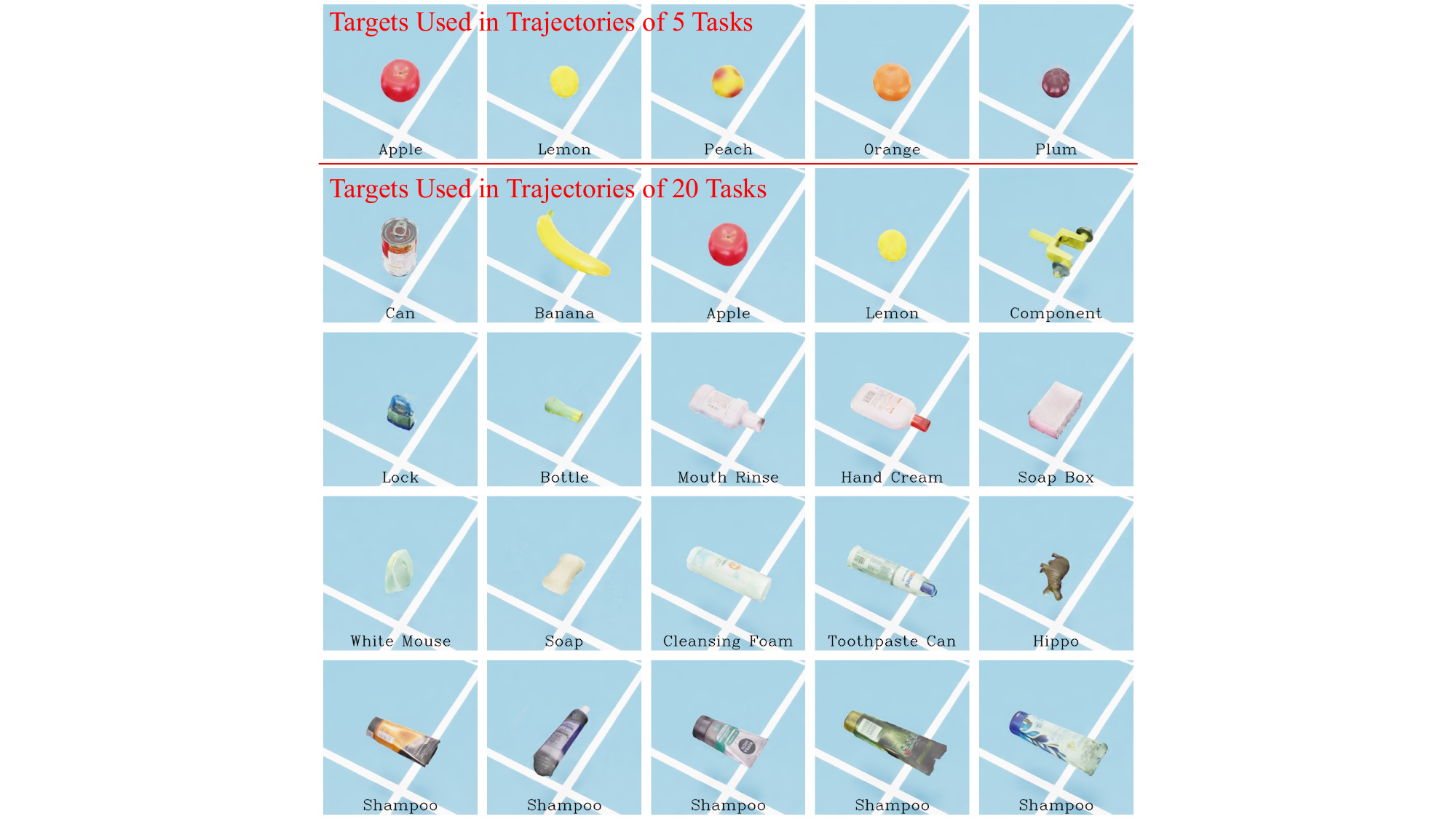}
    \caption{Target objects used in our experiments.}
    \label{fig:objs}
\end{figure}

Our dataset are collected in the simulation environment of IsaacSim, which is automatically collected using algorithm similar with Zheng \etal~\cite{zheng2024robocas}. 
Each scene is initialized with 4 to 6 objects placed randomly on a table, both in position and orientation. A Franka-Panda 7-DoF robotic arm equipped with a two-finger gripper is initialized with a random end-effector pose. A static camera positioned in front of the table, along with a wrist camera mounted on the gripper, are used to capture the RGB and depth observation of the scene, as shown in \cref{fig:motivation}(b). During the collection process, a target is sampled from the objects on the table and assigned as the target, and a language instruction is generated using pre-defined templates. At each step, the pose of the gripper, action targets generated by the algorithm, robot joint information, gripper status, images from the cameras, task instruction, and status information of all objects in the scene, are recorded for training and reproducing. We use ray-tracing renderer while generating camera images and evaluation. The target objects used in our experiments are shown in ~\cref{fig:objs}.

For full-stage trajectories, we first sample collision-free grasp labels on the target, which is densely labeled using the collision model of the object using method of Fang \etal~\cite{fang2020graspnet}. The agent then performs 6-DoF path planning using CuRobo~\cite{curobo_icra23} and executes the generated path. 
For trajectories that only involve \srp stage, to provide a implementable pipeline in real-world robots, we did not directly read the target information from the simulator. Following the method provided in \cref{fig:motivation}(c), we first located the target from the RGB image captured by the static camera, after which it is feed to a CNN to detect the target bounding box. With the bounding box, we can acquire the mean depth of the target from the depth image, and calculate its position using the intrinsic and extrinsic of the camera.
Then in the approach pose sampling stage, we simply samples an end-effector pose within a range of $10cm$ from the target position, ensuring the gripper is oriented towards the target, after which the paths are planned with spatial occupancy information provided by the depth image, and finally the path is executed by the agent.

\subsection{CALVIN Dataset}
\label{sec:calvin}

In our datasets we use 5 tasks in D-D split of CALVIN dataset, and the tasks are shown in \cref{tab:calvin_tasks}. In the dataset, each trajectory is divided into \srp and \pip segments using the methods outlined in \cref{sec:data-mixture}. In the experiments presented in \cref{tab:diff_models}, we use a total of 150 trajectories per task, setting the parameter $p_{\srp} = 50\%$ for the model trained with our method.

\begin{table}
  \centering
  \begin{tabular}{@{}cl@{}}
    \toprule
    \# & Task \\
    \midrule
    1 & lift\_blue\_block\_table \\
    2 & lift\_red\_block\_table \\
    3 & lift\_pink\_block\_table \\
    4 & move\_slider\_left \\
    5 & move\_slider\_right \\
    \bottomrule
  \end{tabular}
  \caption{Tasks we used in the CALVIN dataset.}
  \label{tab:calvin_tasks}
\end{table}
\section{Experiment Details}

\subsection{Environment Setup}
\label{sec:env-setup}

\paragraph{VLA model}
Unless otherwise stated, in this paper the RoboMM~\cite{yan2024robomm} is used as our baseline, which is a multi-modal VLA model that utilizes UVFormer~\cite{liu2024robouniview} to help with spatial perception through RGB image with camera parameters in a low-cost manner. During training we feed language instruction and RGB images from a static camera and a wrist camera, together their intrinsic and extrinsic parameters, into the model, and use the depth images with action chunks as supervision.

\paragraph{Training data}
In the simulation environment of IsaacSim, we generated a dataset for object-picking tasks involving target objects of various categories and geometrical shapes. 
For \srp-only trajectories, to provide a implementable pipeline in real-world robots, we did not directly read the object information from simulation, instead we applied a detection-sampling method provided in \cref{fig:motivation}(c). Details are shown in \cref{sec:collection}.
During trajectory generation, we observe that the \srp-only trajectories are generated $2.5 \times$ faster than those using full-stage data, while the length of the full-stage data is only $1.4 \times$ that of the \srp data. In real-world data collection this discrepancy can only be even larger. 
For the CALVIN dataset used in the experiment, trajectory segments are divided based on the distance and direction of the gripper. Specifically, assume the gripper approach direction vector is $\mathbf{a}_n$ (as shown in \cref{fig:grasp-annotation}) and the vector from the gripper to the target object is $\mathbf{v}$, before the gripper manipulates the target object, the trajectory segment is labeled as \pip if $\left \| \mathbf{v} \right \| \le 0.2m$ and $\left \langle \mathbf{a}_n, \mathbf{v}  \right \rangle \le \frac{\pi}{3}$, and the segment before \pip is labeled as \srp. Details are shown in \cref{sec:calvin}.

\paragraph{Evaluation}
We evaluated our models in the aforementioned simulation environment. A trail is considered succeeded if the agent successfully picked up the instruction-specified target object under the actions generated by the VLA model. During evaluation the scenes are divided into test and zero-shot configurations. The test scenes are configured in ways that have been encountered during training, while the zero-shot scenes are initialized randomly to test the generalization performance. Note that depth images are not utilized in the model inference, only language instruction, RGB observation and camera parameters are fed into the model.

\paragraph{Training}
Our models are trained on servers equipped with 8 Nvidia A100 GPUs, each with 80GB of CUDA memory. The \srp segments are generally longer than the \pip segments, and the dataset $\mathcal{D^{Mix}}$ contains several times more \srp trajectories compared to \pip trajectories. During training we form the mixed dataset $\mathcal{D}^{Mix}$ with a varying of proportion of independent \srp segments, \ie
\begin{equation}
    p_{\srp} = \frac{N_2}{N_1 + N_2}.
\end{equation}
To prevent the \srp features from dominating the model's understanding of operations, during training, the \pip trajectories $\tau^{\pip} \subset \tau^F$ are duplicated $\left \lfloor \frac{N_2}{N_1} \right \rfloor$ times. The benefits of this approach are discussed in \cref{sec:balancing}. 
The checkpoint with the best performance in zero-shot environments within the first 10 epochs is used for evaluation.

\subsection{Scaling with \srp Proportion}
\label{sec:proportion}
\begin{figure}[t]
  \centering
  \includegraphics[width=\linewidth, trim=2 2 2 2, clip]{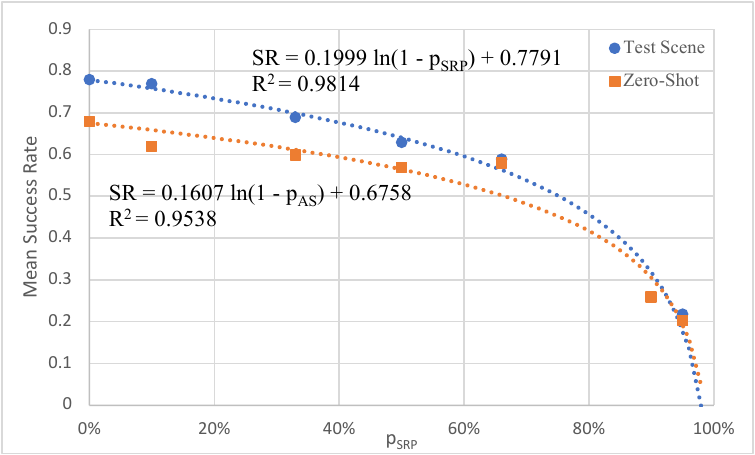}
  \caption{Model performance with the proportion of \srp data. All models are trained with the same amount of total trajectories.}
  \label{fig:percentage}
\end{figure}

To determine the proportion of inexpensive \srp trajectories $p_{\srp}$ that can be incorporated into the model without significantly impacting performance, we trained several models on 5 tasks, each with a total of $N_1 + N_2 = 300$ trajectories. The results are presented in \cref{fig:percentage}. It can be observed that there is a logarithmic relationship between the the proportion of \srp data and the model success rate, expressed as $SR = k \ln \left(1 - p_{\srp}\right) + b$, and when $P_{\srp} > 66\%$, the rate of decline in mean success rate increases rapidly. This decline is due to the increasing reliance on trajectories that lack a manipulation phase, which undermines the model's object manipulation skills, causing the agent to merely wander around the target. In particular, when $\frac{d RS}{d p_{\srp}} = \frac{k}{p_{\srp} - 1} = -1$, \ie $p_{\srp} = 1 - k \approx 80\%$ in our experiments, the rate of performance decline exceeds that of increasing the proportion of independent \srp data. This result suggests us that for a existing dataset $\mathcal{D}^{F}$ containing $N_1$ full-stage trajectories, we can add at most $N_2 = \left|\mathcal{D}^{\srp}_{ind}\right| = 4N_1$ additional independent approaching trajectories into $\mathcal{D}^F$ to form the mixture dataset $\mathcal{D}^{Mix}$ according to \cref{equ:mix-dataset} during model training, to maximize the contribution of the expensive full-stage data.

\subsection{Additional Experiments}

\begin{table}
  \centering
  \begin{tabular}{@{}c|cc@{}}
    \toprule
    $N_2$ & Seen Targets & Novel Targets \\
    \midrule
    50 & 0.235 & 0.2 \\
    100 & 0.375 & 0.275 \\
    200 & 0.3467 & 0.3025 \\
    \bottomrule
  \end{tabular}
  \caption{Generalization performance of models trained with additional data of novel targets. We set $N_1 = 100, N_2=0$ for seen targets and $N_1 = 50$ for novel targets.}
  \label{tab:half_traj_20t}
\end{table}

To assess the feasibility of integrating data on novel target objects into existing comprehensive datasets collected with substantial effort, we trained models using a dataset consisting of 10 novel objects, to serve as a supplementary experiment to the results shown in \cref{tab:half_trajs}. This dataset included a limited amount of full-stage data and a large proportion of \srp data, complemented by another 10 target objects with complete full-stage data. The results in zero-shot scenes are shown in \cref{tab:half_traj_20t}, and the detailed result is shown in \cref{tab:half_traj_20t-detail}. A similar conclusion can be drawn as demonstrated in \cref{fig:sr_wrt_trajs}: by incorporating more \srp data along with a small proportion of full-stage trajectories, we can enhance performance on novel targets.

\begin{table*}[t]
  \centering
  \begin{tabular}{@{}ll|cc|cc|cc@{}}
    \toprule
    & \multirow{2}{*}{Target Object} & \multicolumn{2}{c|}{$N_2=50$} & \multicolumn{2}{c|}{$N_2=100$} & \multicolumn{2}{c}{$N_2=200$} \\
    & & TS & ZS & TS & ZS & TS & ZS \\
    \midrule
    \multirow{11}{*}{\rotatebox{90}{\makecell[c]{Seen Targets\\($N_1 = 100, N_2 = 0$)}}} & Apple & 0.6 & 0.2 & 0.75 & 0.45 & 0.7 & 0.5 \\
    & Banana & 0.75 & 0.2 & 0.8 & 0.35 & 0.8 & 0.15 \\
    & Bottle & 0.75 & 0.15 & 0.8 & 0.45 & 0.85 & 0.3 \\
    & Can & 0.8 & 0.4 & 0.8 & 0.35 & 0.65 & 0.35 \\
    & Toothpaste Can & 0.55 & 0.25 & 0.6 & 0.35 & 0.65 & 0.2 \\
    & Component & 0.655 & 0.4 & 0.75 & 0.5 & 0.6 & 0.4167 \\
    & Hippo & 0.45 & 0.15 & 0.6 & 0.35 & 0.5 & 0.5 \\
    & Lock & 0.75 & 0.25 & 0.75 & 0.65 & 0.75 & 0.55 \\
    & Soap Box & 0.4 & 0.1 & 0.4 & 0.1 & 0.2 & 0.25 \\
    & White Mouse & 0.35 & 0.25 & 0.45 & 0.2 & 0.6 & 0.25 \\
    \cmidrule{2-8}
    & \textbf{Average} & 0.6055 & 0.235 & 0.67 & 0.375 & 0.63 & 0.3467 \\
    \midrule
    \multirow{11}{*}{\rotatebox{90}{\makecell[c]{Novel Targets\\($N_1 = 50$)}}} & Cleansing Foam & 0.65 & 0 & 0.35 & 0.15 & 0.25 & 0.25 \\
    & Hand Cream & 0.4 & 0.15 & 0.6 & 0.2 & 0.4 & 0.2 \\
    & Lemon & 0.65 & 0.35 & 0.7 & 0.45 & 0.85 & 0.45 \\
    & Mouth Rinse & 0.5 & 0.2 & 0.75 & 0.25 & 0.55 & 0.3 \\
    & Shampoo (1) & 0.6 & 0.2 & 0.55 & 0.25 & 0.7 & 0.225 \\
    & Shampoo (2) & 0.65 & 0.2 & 0.55 & 0.3 & 0.5 & 0.35 \\
    & Shampoo (3) & 0.5 & 0.2 & 0.65 & 0.3 & 0.65 & 0.325 \\
    & Shampoo (4) & 0.5 & 0.2 & 0.6 & 0.15 & 0.45 & 0.2 \\
    & Shampoo (5) & 0.55 & 0.3 & 0.7 & 0.4 & 0.5 & 0.475 \\
    & Soap & 0.55 & 0.2 & 0.5 & 0.3 & 0.4 & 0.25 \\
    \cmidrule{2-8}
    & \textbf{Average} & 0.555 & 0.2 & 0.595 & 0.275 & 0.525 & 0.3025 \\
    \bottomrule
  \end{tabular}
  \caption{Detailed results of models trained with additional data of novel targets.}
  \label{tab:half_traj_20t-detail}
\end{table*}

\subsection{Detailed Experiment Results}
The success rate of each task in each model presented in \cref{sec:exps} are shown from \cref{tab:upper_bound-detail-5t} to \cref{tab:half_traj-details}.

\begin{table*}[t]
  \centering
  \begin{tabular}{@{}l|cc|cc|cc|cc|cc|cc@{}}
    \toprule
    \multirow{2}{*}{Target Object} & \multicolumn{2}{c|}{$N_1=50$} & \multicolumn{2}{c|}{$N_1=100$} & \multicolumn{2}{c|}{$N_1=200$} & \multicolumn{2}{c|}{$N_1=300$} & \multicolumn{2}{c|}{$N_1=400$} & \multicolumn{2}{c}{$N_1=500$} \\
    & TS & ZS & TS & ZS & TS & ZS & TS & ZS & TS & ZS & TS & ZS \\
    \midrule
    Plum & 0.5 & 0.4 & 0.35 & 0.75 & 0.75 & 0.7 & 0.8 & 0.9 & 0.8 & 0.9 & 0.95 & 0.9 \\
    Lemon & 0.5 & 0.15 & 0.55 & 0.05 & 0.65 & 0.25 & 0.75 & 0.55 & 0.846 & 0.875 & 0.65 & 0.85 \\
    Orange & 0.35 & 0 & 0.25 & 0.15 & 0.85 & 0.25 & 0.9 & 0.4 & 0.9 & 0.6 & 0.9 & 0.7 \\
    Apple & 0.425 & 0.1 & 0.25 & 0.2 & 0.75 & 0.5 & 0.8 & 0.75 & 0.95 & 0.8 & 0.85 & 0.75 \\
    Peach & 0.4 & 0.15 & 0.45 & 0.25 & 0.55 & 0.45 & 0.65 & 0.8 & 0.8 & 0.85 & 0.8 & 0.75 \\
    \midrule
    \textbf{Average} & 0.435 & 0.16 & 0.37 & 0.28 & 0.71 & 0.43 & 0.78 & 0.68 & 0.8592 & 0.805 & 0.83 & 0.79 \\
    \bottomrule
  \end{tabular}
  \caption{Detailed results of upper bound model performance used in \cref{fig:sr_wrt_trajs}. (5 Tasks, $N_2=0$)}
  \label{tab:upper_bound-detail-5t}
\end{table*}

\begin{table*}
  \centering
  \begin{tabular}{@{}l|cc|cc|cc|cc@{}}
    \toprule
    \multirow{2}{*}{Target Object} & \multicolumn{2}{c|}{$N_1=100$} & \multicolumn{2}{c|}{$N_1=200$} & \multicolumn{2}{c|}{$N_1=300$} & \multicolumn{2}{c}{$N_1=400$} \\
    & TS & ZS & TS & ZS & TS & ZS & TS & ZS \\
    \midrule
    Apple & 0.7 & 0.35 & 0.8 & 0.5 & 0.85 & 0.55 & 0.85 & 0.55 \\
    Banana & 0.8 & 0.35 & 0.7 & 0.4 & 0.65 & 0.45 & 0.8 & 0.85 \\
    Bottle & 0.9 & 0.3 & 0.7 & 0.45 & 0.8 & 0.5 & 0.7179 & 0.6 \\
    Can & 0.6 & 0.3 & 0.8 & 0.65 & 0.8 & 0.55 & 0.7 & 0.6 \\
    Toothpaste Can & 0.7 & 0.35 & 0.65 & 0.4 & 0.65 & 0.4 & 0.6 & 0.45 \\
    Cleansing Foam & 0.8 & 0.15 & 0.55 & 0.15 & 0.75 & 0.4 & 0.85 & 0.6 \\
    Component & 0.864 & 0.35 & 0.65 & 0.4 & 0.75 & 0.65 & 0.75 & 0.45 \\
    Hand Cream & 0.55 & 0.2 & 0.5 & 0.45 & 0.7 & 0.4 & 0.8 & 0.4 \\
    Hippo & 0.6 & 0.275 & 0.7 & 0.35 & 0.75 & 0.55 & 0.65 & 0.75 \\
    Lemon & 0.7 & 0.4 & 0.9 & 0.6 & 0.95 & 0.6 & 0.9 & 0.85 \\
    Lock & 0.65 & 0.3 & 0.75 & 0.4 & 0.9 & 0.7 & 0.7 & 0.8 \\
    Mouth Rinse & 0.8 & 0.4 & 0.6 & 0.5 & 0.9 & 0.5 & 0.9 & 0.5 \\
    Shampoo (1) & 0.8 & 0.25 & 0.55 & 0.4 & 0.7 & 0.5 & 0.75 & 0.55 \\
    Shampoo (2) & 0.711 & 0.1 & 0.7037 & 0.4 & 0.6 & 0.4554 & 0.6 & 0.55 \\
    Shampoo (3) & 0.6 & 0.3 & 0.7 & 0.425 & 0.65 & 0.45 & 0.85 & 0.375 \\
    Shampoo (4) & 0.7 & 0.2 & 0.7 & 0.425 & 0.8 & 0.5 & 0.65 & 0.55 \\
    Shampoo (5) & 0.75 & 0.35 & 0.65 & 0.7 & 0.55 & 0.65 & 0.6 & 0.6 \\
    Soap & 0.85 & 0.3 & 0.9 & 0.6 & 0.8 & 0.55 & 0.8696 & 0.4 \\
    Soap Box & 0.25 & 0.2 & 0.4872 & 0.25 & 0.45 & 0.35 & 0.5 & 0.45 \\
    White Mouse & 0.5 & 0.25 & 0.65 & 0.3 & 0.7 & 0.45 & 0.75 & 0.35 \\
    \midrule
    \textbf{Average} & 0.6913 & 0.2838 & 0.682 & 0.4375 & 0.735 & 0.5128 & 0.7394 & 0.5413 \\
    \bottomrule
  \end{tabular}
  \caption{Detailed results of upper bound model performance used in \cref{fig:sr_wrt_trajs}. (20 Tasks, $N_2=0$)}
  \label{tab:upper_bound-detail-20t}
\end{table*}

\begin{table*}
  \centering
  \begin{tabular}{@{}ll|cc|cc|cc|cc@{}}
    \toprule
    & \multirow{2}{*}{Target Object} & \multicolumn{2}{c|}{$N_2=100$} & \multicolumn{2}{c|}{$N_2=200$} & \multicolumn{2}{c|}{$N_2=300$} & \multicolumn{2}{c}{$N_2=400$} \\
    & & TS & ZS & TS & ZS & TS & ZS & TS & ZS \\
    \midrule
    \multirow{6}{*}{\rotatebox{90}{5 Tasks}} & Plum & 0.65 & 0.55 & 0.65 & 0.8 & 0.6 & 0.75 & 0.7 & 0.75 \\
    & Lemon & 0.6 & 0.1 & 0.65 & 0.55 & 0.6625 & 0.85 & 0.685 & 0.685 \\
    & Orange & 0.55 & 0.5 & 0.7 & 0.35 & 0.6 & 0.55 & 0.65 & 0.55 \\
    & Apple & 0.5 & 0.2 & 0.65 & 0.5 & 0.7 & 0.6 & 0.85 & 0.7 \\
    & Peach & 0.5 & 0.5 & 0.3 & 0.7 & 0.67 & 0.6 & 0.575 & 0.7 \\
    \cmidrule{2-10}
    & \textbf{Average} & 0.56 & 0.37 & 0.59 & 0.58 & 0.6925 & 0.67 & 0.712 & 0.695 \\
    \midrule
    \multirow{21}{*}{\rotatebox{90}{20 Tasks}} & Apple & 0.55 & 0.2 & 0.55 & 0.5 & 0.4 & 0.4 & - & - \\
    & Banana & 0.75 & 0.2 & 0.75 & 0.325 & 0.65 & 0.55 & - & - \\
    & Bottle & 0.65 & 0.25 & 0.775 & 0.55 & 0.7 & 0.3 & - & - \\
    & Can & 0.6 & 0.15 & 0.6 & 0.4667 & 0.5 & 0.45 & - & - \\
    & Toothpaste Can & 0.6 & 0.2 & 0.5 & 0.35 & 0.5 & 0.15 & - & - \\
    & Cleansing Foam & 0.5 & 0.25 & 0.65 & 0.1 & 0.5 & 0.6 & - & - \\
    & Component & 0.65 & 0.1 & 0.7 & 0.45 & 0.65 & 0.25 & - & - \\
    & Hand Cream & 0.5 & 0.45 & 0.75 & 0.2 & 0.45 & 0.25 & - & - \\
    & Hippo & 0.65 & 0.35 & 0.55 & 0.5 & 0.55 & 0.75 & - & - \\
    & Lemon & 0.65 & 0.4 & 0.95 & 0.55 & 0.75 & 0.75 & - & - \\
    & Lock & 0.35 & 0.35 & 0.85 & 0.7 & 0.45 & 0.45 & - & - \\
    & Mouth Rinse & 0.65 & 0.3 & 0.55 & 0.45 & 0.75 & 0.35 & - & - \\
    & Shampoo (1) & 0.6 & 0.2 & 0.65 & 0.45 & 0.55 & 0.4 & - & - \\
    & Shampoo (2) & 0.7 & 0.35 & 0.7 & 0.3448 & 0.7 & 0.45 & - & - \\
    & Shampoo (3) & 0.55 & 0.4 & 0.7027 & 0.35 & 0.7 & 0.5 & - & - \\
    & Shampoo (4) & 0.55 & 0.15 & 0.6 & 0.5 & 0.5 & 0.35 & - & - \\
    & Shampoo (5) & 0.7 & 0.55 & 0.65 & 0.55 & 0.65 & 0.65 & - & - \\
    & Soap & 0.5 & 0.3 & 0.65 & 0.4 & 0.6 & 0.45 & - & - \\
    & Soap Box & 0.25 & 0.15 & 0.4 & 0.3 & 0.6 & 0.3 & - & - \\
    & White Mouse & 0.4 & 0.2 & 0.65 & 0.25 & 0.35 & 0.25 & - & - \\
    \cmidrule{2-10}
    & \textbf{Average} & 0.7675 & 0.275 & 0.6589 & 0.4143 & 0.575 & 0.43 & - & - \\
    \bottomrule
  \end{tabular}
  \caption{Detailed results of models trained with different amount of \srp data used in \cref{fig:sr_wrt_trajs}. ($N_1=100$)}
  \label{tab:upper_bound-detail}
\end{table*}

\begin{table*}
  \centering
  \begin{tabular}{@{}l|cc|cc|cc@{}}
    \toprule
    \multirow{2}{*}{Target Object} & \multicolumn{2}{c|}{\makecell{w/o \srp\\($N_1=100, N_2=0$)}} & \multicolumn{2}{c|}{\makecell{w/ \srp\\($N_1=100, N_2=200$)}} & \multicolumn{2}{c}{\makecell{UB\\($N_1=300, N_2=0$)}} \\
    & TS & ZS & TS & ZS & TS & ZS \\
    \midrule
    Plum & 0.3 & 0.25 & 0.35 & 0.35 & 0.85 & 0.7 \\
    Lemon & 0.1364 & 0.05 & 0.45 & 0.35 & 0.6 & 0.35 \\
    Orange & 0.15 & 0.05 & 0.3 & 0.1 & 0.3 & 0.2 \\
    Apple & 0.15 & 0.1 & 0.35 & 0.1 & 0.55 & 0.4 \\
    Peach & 0.2 & 0.2 & 0.35 & 0.5 & 0.5 & 0.55 \\
    \midrule
    \textbf{Average} & 0.1873 & 0.13 & 0.36 & 0.28 & 0.56 & 0.44 \\
    \bottomrule
  \end{tabular}
  \caption{Detailed results of \cref{tab:diff_models}. Performance of RoboFlamingo~\cite{li2023vision} on our dataset.}
  \label{tab:rf-ours}
\end{table*}

\begin{table*}
  \centering
  \begin{tabular}{@{}l|cc|cc|cc@{}}
    \toprule
    \multirow{2}{*}{Target Object} & \multicolumn{2}{c|}{\makecell{w/o \srp\\($N_1=75, N_2=0$)}} & \multicolumn{2}{c|}{\makecell{w/ \srp\\($N_1=75, N_2=75$)}} & \multicolumn{2}{c}{\makecell{UB\\($N_1=150, N_2=0$)}} \\
    & RM & RF & RM & RF & RM & RF \\
    \midrule
    lift\_blue\_block\_table & 0.4583 & 0.3333 & 0.5833 & 0.2917 & 0.7717& 0.5 \\
    lift\_pink\_block\_table & 0.6 & 0.16 & 0.92 & 0.48 & 0.88 & 0.6 \\
    lift\_red\_block\_table & 0.3333 & 0.0833 & 0.625 & 0.2083 & 0.9183& 0.1667 \\
    move\_slider\_left & 0.9219 & 1 & 0.9219 & 1 & 0.9244& 1 \\
    move\_slider\_right & 1 & 1 & 1 & 1 & 1 & 1 \\
    \midrule
    \textbf{Average} & 0.8036 & 0.7366 & 0.8839 & 0.7812 & 0.9254& 0.8125 \\
    \bottomrule
  \end{tabular}
  \caption{Detailed results of \cref{tab:diff_models}. Performance of RoboMM (RM)~\cite{yan2024robomm} and RoboFlamingo (RF)~\cite{li2023vision} on CALVIN.}
  \label{tab:models-calvin}
\end{table*}

\begin{table*}
  \centering
  \begin{tabular}{@{}ll|ccc@{}}
    \toprule
    \makecell{Object\\Group} & Target Object & \makecell{None\\($N_1=0, N_2=0$)} & \makecell{Only \srp\\($N_1=0, N_2=200$)} & \makecell{Full-Stage\\($N_1=200, N_2=0$)} \\
    \midrule
    \multirow{5}{*}{\makecell[l]{Seen\\$N_1=200$\\$N_2=0$}} & Apple & 0.65 & 0.5 & 0.5 \\
    & Lemon & 0.65 & 0.7 & 0.6 \\
    & Orange & 0.75 & 0.7 & 0.65 \\
    & Peach & 0.65 & 0.7 & 0.8 \\
    \cmidrule{2-5}
    & \textbf{Average} & 0.675 & 0.65 & 0.6375 \\
    \midrule
    Novel & Plum & 0.05 & 0.4 & 0.65 \\
    \bottomrule
  \end{tabular}
  \caption{Detailed results of \cref{tab:half_trajs}. Performance of the models on seen and novel target objects in test scenes. (5 Tasks)}
  \label{tab:half_traj-details-5t}
\end{table*}

\begin{table*}
  \centering
  \begin{tabular}{@{}ll|ccc@{}}
    \toprule
    \makecell{Object\\Group} & Target Object & \makecell{None\\($N_1=0, N_2=0$)} & \makecell{Only \srp\\($N_1=0, N_2=100$)} & \makecell{Full-Stage\\($N_1=100, N_2=0$)} \\
    \midrule
    \multirow{11}{*}{\makecell[l]{Seen\\$N_1=100$\\$N_2=0$}} & Apple & 0.35 & 0.4 & 0.7 \\
    & Banana & 0.6 & 0.8 & 0.8 \\
    & Bottle & 0.65 & 0.825 & 0.9 \\
    & Can & 0.7 & 0.65 & 0.6 \\
    & Toothpaste Can & 0.65 & 0.65 & 0.7 \\
    & Cleansing Foam & 0.6 & 0.55 & 0.8 \\
    & Component & 0.7 & 0.5 & 0.864 \\
    & Hippo & 0.5 & 0.5 & 0.6 \\
    & Shampoo (2) & 0.7 & 0.7 & 0.711 \\
    & Shampoo (4) & 0.6 & 0.5 & 0.7 \\
    \cmidrule{2-5}
    & \textbf{Average} & 0.605 & 0.6075 & 0.7375 \\
    \midrule
    \multirow{11}{*}{Novel} & Hand Cream & 0 & 0.2 & 0.55 \\
    & lemon & 0.05 & 0.225 & 0.7 \\
    & Lock & 0.05 & 0.15 & 0.65 \\
    & Mouth Rinse & 0 & 0.3 & 0.8 \\
    & Shampoo (1) & 0 & 0.15 & 0.8 \\
    & Shampoo (3) & 0.05 & 0.175 & 0.6 \\
    & Shampoo (5) & 0.05 & 0.45 & 0.75 \\
    & Soap & 0 & 0.075 & 0.85 \\
    & Soap Box & 0 & 0.15 & 0.25 \\
    & White Mouse & 0.05 & 0.15 & 0.5 \\
    \cmidrule{2-5}
    & \textbf{Average} & 0.025 & 0.2025 & 0.645 \\
    \bottomrule
  \end{tabular}
  \caption{Detailed results of \cref{tab:half_trajs}. Performance of the models on seen and novel target objects in test scenes. (20 Tasks)}
  \label{tab:half_traj-details}
\end{table*}

\subsection{Failure Analysis}
\begin{figure}
    \centering
    \includegraphics[width=0.8\linewidth, trim=60 20 60 5, clip]{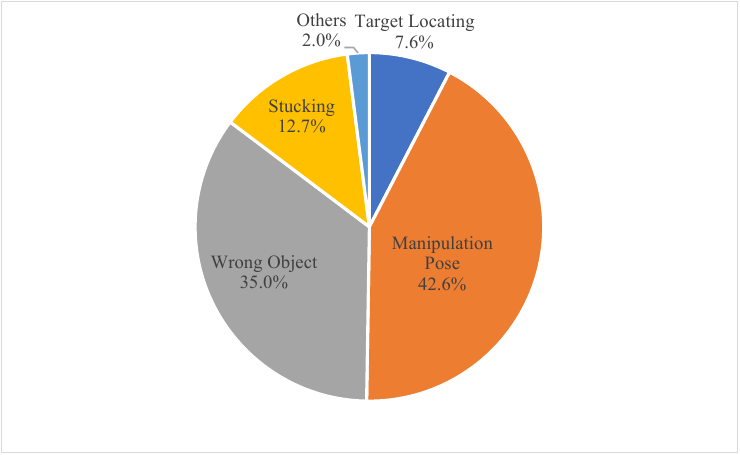}
    \caption{
        Statistical analysis of failure reasons in our experiments. Most failure cases are cause by inaccuracies in manipulation poses, manipulation of the wrong objects, robot getting stuck due to obstacles or kinematic limitations, and failures in locating target object.
    }
    \label{fig:failure-analysis}
\end{figure}

During the evaluation process of the policies, we observed that the primary causes of trail failures were inaccuracies in action prediction/execution (42.6\%), as well as incorrect target identification (35\%), as shown in \cref{fig:failure-analysis}. Addressing the latter issue can be achieved through VLM pretraining using massive Internet-scale datasets,  which is beyond the scope of this paper. However, resolving the former challenge is considerably more complex as it involves not only model design but also the development of platforms for data collection and policy evaluation. According to our observation, during the picking phase, even a slight difference of only 0.5$cm$ in pick positions can lead to completely opposite outcomes. While for the trails that are successfully picked the target, only 11.9\% failed during the dropping phase, which does not require high accuracy, mainly due to reaching the maximum step constraint as a result of the time-consuming picking phase. Given that our \projname platform achieves an execution accuracy of 5\% in a single step, which significantly more accurate than other platforms, designing a policy that emphasizes the impact of geometric observations on precise agent actions will be a crucial area for future work.

\end{document}